\documentclass{article}

     \PassOptionsToPackage{numbers, compress}{natbib}

     \usepackage[preprint]{neurips_2021}

\usepackage[utf8]{inputenc} 
\usepackage[T1]{fontenc}    
\usepackage{url}            
\usepackage{booktabs}       
\usepackage{amsfonts}       
\usepackage{nicefrac}       
\usepackage{microtype}      
\usepackage{xcolor}         

\usepackage{times}
\usepackage{epsfig}
\usepackage{graphicx}
\usepackage{amsmath}
\usepackage{amssymb}
\usepackage{multirow}
\usepackage{multicol}
\usepackage{bbding}
\usepackage{subfig}
\usepackage{booktabs}

\title{Coarse-To-Fine Incremental Few-Shot Learning}

\author{Xiang Xiang, Yuwen Tan, Qian Wan, Jing Ma\\
School of Airtificial Intelligence and Automation \\
Huazhong University of Science and Technology, China\\
{\tt\small xex@hust.edu.cn}
}

\begin{document}

\maketitle

\begin{abstract}
  Different from fine-tuning models pre-trained on a large-scale dataset of preset classes,  class-incremental learning (CIL) aims to recognize novel classes over time without forgetting pre-trained classes. However, a given model will be challenged by test images with finer-grained classes, e.g., a basenji is at most recognized as a dog. Such images form a new training set (i.e., support set) so that the incremental model is hoped to recognize a basenji (i.e., query) as a basenji next time. This paper formulates such a hybrid natural problem of coarse-to-fine few-shot (C2FS) recognition as a CIL problem named C2FSCIL, and proposes a simple, effective, and theoretically-sound strategy Knowe: to learn, normalize, and freeze a classifier's weights from fine labels, once learning an embedding space contrastively from coarse labels. Besides, as CIL aims at a stability–plasticity balance, new overall performance metrics are proposed. In that sense, on CIFAR-100, BREEDS, and tieredImageNet, Knowe outperforms all recent relevant CIL/FSCIL methods that are tailored to the new problem setting for the first time.
\end{abstract}

\section{Introduction}
Product visual search is normally driven by a deep model pre-trained on a large-scale private image-set, while at inference it needs to recognize consumer images at a finer granularity.
Such a model is expected to evolve on-the-fly \cite{mai21survey} over time as being used, because fine-tuning (FT) it for specific novel classes induces an increasing number of separate models retrained, and thus is inefficient. This expectation is also generally valid for vision-driven autonomous systems or intelligent agents. For example, a self-driving car needs to gradually grow its perception capabilities as it runs on the road.

\begin{figure}
    \centering
    \includegraphics[scale=0.36]{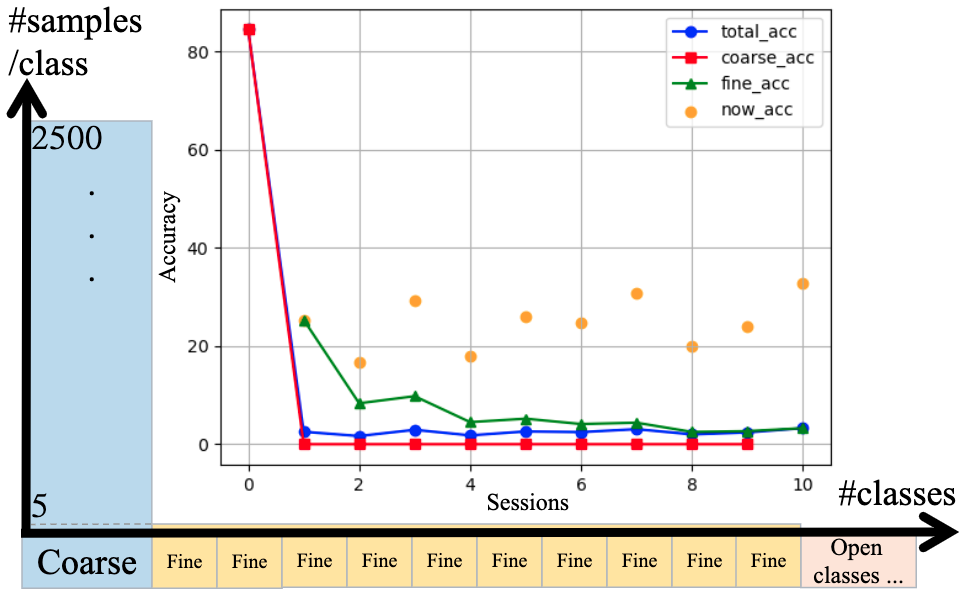}
    \caption{Catastrophic forgetting when FT-ing a coarsely-trained model on fine samples presently available w/o freezing any weight. We pre-set 10 sessions from CIFAR-100 \cite{cifar}. There is a fine-class accuracy from the 1st session and yet no coarse-class accuracy as all samples are with fine labels.}
    \label{fig:cata}
\end{figure}


As shown in Fig. \ref{fig:cata}, we are interested in such a coarse-to-fine recognition problem that fits the class-incremental learning (CIL) setting. Moreover, fine classes appear asynchronously, which again fits CIL. It is also a few-shot learning problem, as there is no time to collect abundant samples per new class. We name such an incremental few-shot learning problem Coarse-to-Fine Few-Shot Class-Incremental Learning (C2FSCIL), and aim to propose a method that can evolve a generic model to both avoid catastrophic forgetting of source-blind coarse classes (see also Fig. \ref{fig:cata})  and prevent over-fitting the new few-shot fine-grained classes. However,
\noindent {\bf what exactly is the knowledge?} Incremental learning (IL) is aimed for the learning model to adapt to new data without forgetting its existing knowledge, which is also called catastrophic forgetting, a concept in connectionist networks \cite{cog99,kirkpatrick2017overcoming} -- it occurs when the new weight vector is completely inappropriate as a solution for the originally learned pattern. In deep learning (DL), knowledge distillation (KD) is one of the most effective approaches to IL, while there lacks a consensus about what exactly the knowledge is in deep networks. Will it similarly be the weight vectors?

{\bf Is a coarsely-learned embedding space generalizable?} We aim to achieve a superior performance at both the coarse and fine granularity. However, considering the diversity of fine labels, it is infeasible to train a comprehensive fine-grained model beforehand.
Instead, can a model be trained, using coarsely-labeled samples, to classify finely-labeled samples with accuracy comparable to that of a model trained with fine labels \cite{colt21coarse}?
Our hypothesis is yes; then, the next question is how to pre-train a generalizable base model? How to explore a finer embedding space from coarse labels? Namely, what type of knowledge is useful
for fine classes and how can we learn and preserve them \cite{cha21}?

{\bf Can we balance old knowledge and current learning?} (\emph{a.k.a.}, solving the stability–plasticity dilemma \cite{mermillod2013stability,wang21cvpr}). We aim to remember cues of both the pre-trained base classes and fine classes in the previous few-shot sessions. 
Our hypothesis is yes and our preference is a linear classifier as it is flexible, data in-demanding, and efficient to train as well as simple for derivation. Furthermore, the next question is how a linear classifier can evolve the model effectively with a few shots and yet a balanced performance. As presumed, if the knowledge is weights, then freezing weights retains knowledge while updating weights is evolving the model. 

To answer those questions, we will first fine-tune a coarse model to test our hypothesis. Motivated to make CIL as simple as fine-tuning, our contributions are four-fold.

\begin{enumerate}
    \item We propose a new problem and empirical insights for incrementally learning coarse-to-fine with a few shots.
    \item We propose to {\bf learn, normalize, and freeze} weights, a simple process (Knowe, pronounced as 'now') that can effectively solve the problem once we have a base model contrastively learned from coarse labels.
    \item We theoretically analyze why Knowe is a valid solver.
    \item We propose a way to measure balanced performance.
\end{enumerate}


\section{Related Problems}
\subsection{Catastrophic Forgetting (CF)}
To learn over time (\emph{i.e.}, sequential learning), it is suggested in \cite{cog89,cog99} that neural networks can be limited by catastrophic forgetting (CF) just like Perceptron is unable to solve X-OR. Knowledge forgetting, or called catastrophic forgetting/interference is about a learner's memory (\emph{e.g.}, LSTM) and is a result of the stability–plasticity dilemma regarding how to design a model that is sensitive to, but not radically disrupted by, new input \cite{cog89,cog99}. Often, maintaining plasticity results in forgetting old classes while maintaining stability prevents the model from learning new classes, which may be caused by a single set of shared weights 
\cite{cog99}.
\subsection{Weakly-Supervised Learning}
Judging from the fine-class stage (Fig. \ref{fig:cata} middle to right), if we combine a pre-training set and the support set as a holistic training set, then the few-shot fine-grained recognition using a model pre-trained on coarse samples are similar to the \emph{weakly-supervised learning} and specifically \emph{learning from coarse labels} \cite{cvpr21coarse,colt21coarse,xu21,yang21}, \emph{e.g.}, C2FS \cite{cvpr21coarse}.
Ristel \emph{et. al.} investigates how coarse labels can be used to recognize sub-categories using random forests \cite{riste15} (say, NCM \cite{riste14}).
\subsection{Open-World Recognition} 

Judging from the coarse-class stage  \cite{open15} (see the left side of Fig. \ref{fig:cata}), CIL \cite{iCaRL} can be dated back to the support vector machine \cite{il-svm} and random forest \cite{riste14,riste15}, where a new class can be added as a new node,  and now seen as a progressive case of \emph{continual/lifelong learning} \cite{de21survey,mai21survey}, where CF is a challenge as data are hidden. The topology structure is also favored in DL \cite{Tao_2020_CVPR,tao20eccv}.
\emph{Few-shot learning} (FSL) measures models' ability to quickly adapt to new tasks \cite{rethink20} and has a flavor of CIL considering novel classes in the support set, \emph{e.g.}, DFSL \cite{gidaris18}, IFSL \cite{ren19}, FSCIL \cite{Tao_2020_CVPR, dong21, zhang21}, and so on.
\subsection{Uniqueness of Proposed Problem}
\begin{figure}
    \centering
    \includegraphics[scale=0.6]{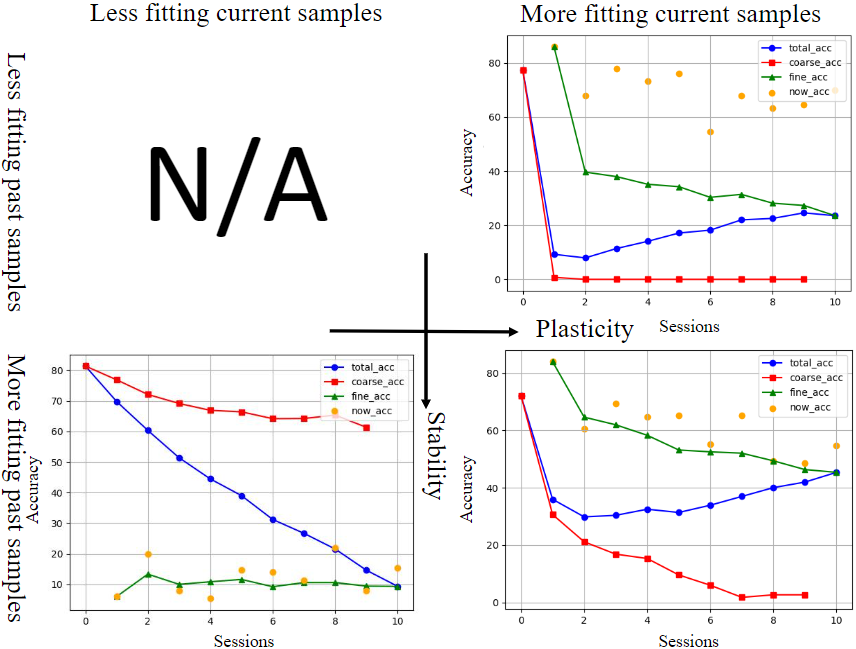}
    \caption{The stability-plasticity trade-off. Top-right is FT w/o IL; bottom-left represents most IFSL methods; bottom-right is our approach; top-left does not apply. (CIFAR-100)}
    \label{fig:diff-sett}
\end{figure}

Different from existing settings \cite{dong21,Tao_2020_CVPR,zhang21} that focus on remembering the pre-trained base classes only, our setting requires remembering the knowledge gained in both the base coarse and previous fine sessions.
We add finer classes instead of new classes at the same granularity.
Our setting requires a balance between coarse and fine performance unexplored by existing works, as shown in Fig. \ref{fig:diff-sett}.

\begin{table}[h]
\resizebox{\linewidth}{!}{
\begin{tabular}{lccc}
\toprule
Method & Class hierarchy & Few-shot Learning & Incremental Learning\\
\midrule
LwF \cite{gidaris18}              & & &\checkmark\\
CEC\cite{zhang21} &   & \checkmark & \checkmark     \\
ANCOR \cite{cvpr21coarse} & \checkmark & \checkmark &    \\
IIRC\cite{abdelsalam2021iirc}&\checkmark  &  & \checkmark   \\
\midrule
C2FSCIL (Ours) & \checkmark & \checkmark & \checkmark     \\
\bottomrule
\end{tabular}
}
\vspace{3mm}
\caption{Comparison of settings with related works.}
\label{tab:table9}
\end{table}

\section{State of the Art (SOTA)}
\subsection{Incremental Learning (IL)}
IL allows a model to be continually updated on new data without forgetting, instead of training a model once on all data. There are two settings: class-IL \cite{marc20survey} and task-IL \cite{de21survey}. 

They share main approaches, such as regularization and rehearsal methods. Regularization methods prevent the drift of consolidated weights and optimize network parameters for the current task, \emph{e.g.}, parameter control in EWC \cite{kirkpatrick2017overcoming}.
CIL is our focus and aims at learning a classifier that maintains a good performance on all classes seen in different sessions. 

In addition, Li \emph{et.al.} first introduces 
KD \cite{hinto15ds} to IL literature in LwF \cite{li2016learning} by modifying a cross-entropy loss to retain the knowledge in the original model. 
Recent works focus on retaining old-class samples to compute the KD loss. For example, iCaRL \cite{iCaRL} learns both features and strong classifiers by combining KD and feature learning, \emph{e.g.}, NME.

\subsection{Operating Weights for IL}
The IL literature since 2017 has seen various weight operations (op. for short) in the sense of consolidation (\emph{e.g.}, EWC \cite{kirkpatrick2017overcoming}), aligning \cite{zhao20cvpr,tale21cvprw}, normalization \cite{zhao20cvpr,zhu21},  standardization \cite{belouadah2020initial}, regularization \cite{kirkpatrick2017overcoming,pan20}, aggregation \cite{liu21}, calibration \cite{cali19nips}, rectification \cite{rec21}, transfer \cite{lee17,liu20cvpr}, sharing \cite{sharing19iclr}, masking \cite{maskwe18}, imprinting \cite{imprint18}, picking \cite{hung19}, scaling \cite{belouadah2020scail}, merging \cite{lee20cvpr}, pruning \cite{pack18cvpr}, 
quantizaton \cite{shi21}, weight importance \cite{jung20}, assignment \cite{hu21}, restricting weights to be positive \cite{zhao20cvpr}, constraining weight changes \cite{anna21}, and so on.


\subsection{Few-Shot Learning (FSL)}

The prosperity of DL has pushed  large-scale supervised learning, so far,  to be the most popular learning paradigm. However, FSL is human-like learning \cite{yao19fsl} in the case of only a few samples \cite{shu2018small,fsl20}. 
For example, Finn \emph{et. al.} proposed Model-Agnostic Meta Learning (MAML) to train a model that can quickly adapt to a new task using only a few samples and training iterations \cite{maml}. Prototypical Network learns
a metric space in which classification can be performed by computing distances to prototype representations of each class \cite{proto}. Ren \emph{et. al.} proposes a meta-learning model, the Attention Attractor Network, which regularizes the learning of novel classes \cite{ren19}. It is shown that decoupling the embedding learner and classifier is feasible \cite{zhang21}. Tian \emph{et. al.} demonstrates that using a good learned embedding model can be more effective than meta learning  \cite{rethink20}.

\subsection{Incremental Few-Shot Learning (IFSL)}

In the IFSL \cite{ren19} or similarly FSCIL \cite{Tao_2020_CVPR} setting, samples in the incremental session are relatively scarce, different from conventional CIL. While IFSL is based on meta learning, IFSL and DFSL \cite{gidaris18} both utilize attentions.
In FSCIL, a model named TOPIC is proposed, which contains a single neural gas (NG) network to learn feature-space topologies as knowledge, and adjust NG to preserve the stabilization and enhance the adaptation. In \cite{dong21}, Dong \emph{et. al.} propose an exemplar relation KD-IL framework to balance the tasks of old-knowledge preserving and new-knowledge adaptation as done in \cite{wu21}. CEC \cite {zhang21} is proposed to separate classifier from the embedding learner, and use a graph attention network to propagate context cues between classifiers for adaptation. In \cite{hou19}, Hou \emph{et. al.} address the imbalance between old and new classes by cosine normalization \cite{normface,gidaris18,hou19}.

\subsection{Uniqueness of Proposed Approach}
Different from state-of-the-art approaches to IFSL, we do not follow rehearsal methods, namely, our model learns without memorizing samples \cite{lwm19}. 
However, retaining samples is often practically infeasible, say, when learning on-the-fly \cite{mai21survey}. Even if there is memory for storing previous samples, there often is a budget, buffer, or queue. 
Thus, we aim to examine the extreme case of knowledge forgetting, and thus design IFSL methods to the upper-bound extent.
For example, although in \cite{anna21} they do not use any base-class training samples and keep the weights of the base classifier frozen, they still use previous samples in their third phase.

\begin{figure}[!t]
    \centering
    \includegraphics[scale=0.55]{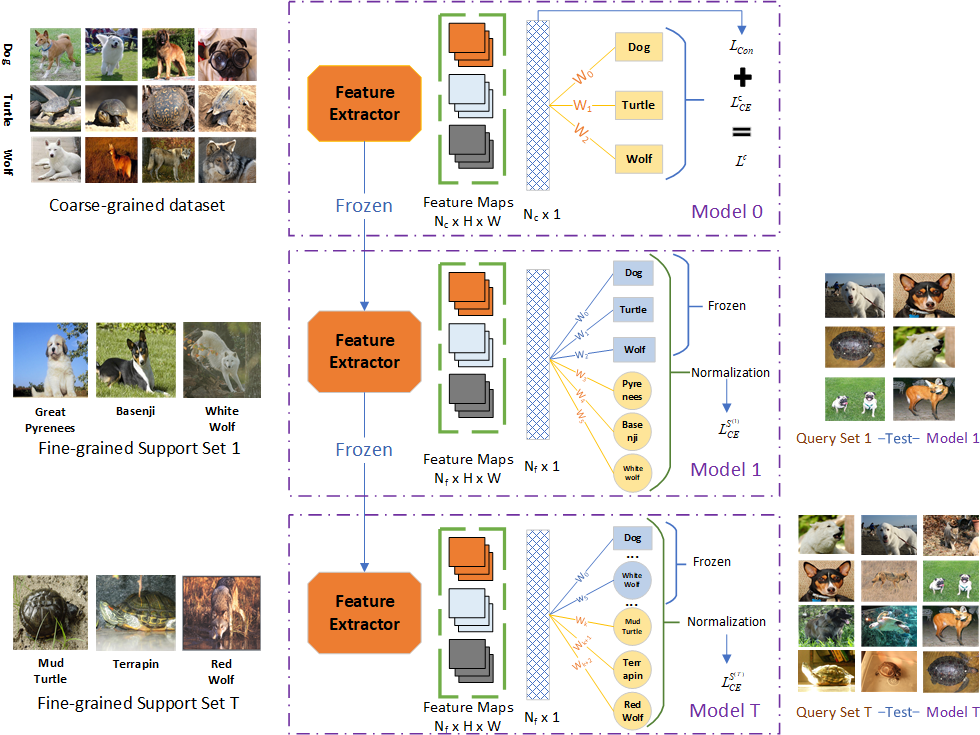}
    \caption{C2FSCIL and basic idea. In base session we train  $\mathbf{\Theta}$ on $\mathbb{D}$ to get $\mathbf{\Theta}^{(0)}$. Per incremental session,  $\mathbf{\Theta}^{(t)}$ is trained on $C$-way $K$-shot support set $\mathbb{S}^{(t)}$ based on $\mathbf{\Theta}^{(t-1)}$, $t \geq 1$ and then tested on any class seen in either $\mathbb{D}$ or $\mathbb{S}^{(1)},...,\mathbb{S}^{(t)}$.}
    \label{fig:proce}
\end{figure}

\section{A New Problem C2FSCIL and Our  Insights}

Given a model parameterized by $\mathbf{\Theta}$ and pre-trained on $\mathbb{D}=\left \{(\mathbf{x}_i,y_i) \right \}_{i=1}^N $ where  $y_i \in \mathbb{Y} = \{\mathcal{Y}_1,\mathcal{Y}_2,...,\mathcal{Y}_R \}$, a set of $R$ coarse labels $\mathcal{Y}$, we have a stream of $C$-way $K$-shot support sets $\mathbb{S}^{(1)},\mathbb{S}^{(2)},...,\mathbb{S}^{(t)},...,\mathbb{S}^{(T)}$ where $\mathbb{S}^{(t)} = \{(\mathbf{x}_j^{(t)},y_j^{(t)}) \}_{j=1}^{C \cdot K}$ and 
$y_j^{(t)} \in \mathbb{Z}^{(t)} = \{\mathcal{Z}^{(t)}_1,...,\mathcal{Z}^{(t)}_C \}$, a set of $C$ fine-grained labels $\mathcal{Z}$. Then, we adapt our model to $\mathbb{S}^{(1)},\mathbb{S}^{(2)},...,\mathbb{S}^{(t)}$ over time and update the parameter set $\mathbf{\Theta}$ from $\mathbf{\Theta}^{(0)}$ all the way to $\mathbf{\Theta}^{(t)}$, as shown levelwise in Fig. \ref{fig:proce}.

For testing, we also have a stream of $(C\cdot t+R)$-way $H$-shot query sets $\mathbb{Q}^{(1)},\mathbb{Q}^{(2)},...,\mathbb{Q}^{(t)},...,\mathbb{Q}^{(T)}$ where $\mathbb{Q}^{(t)} = \{(\mathbf{x}_k^{(t)},y_k^{(t)}) \}_{k=1}^{(Ct+R)H}$ and 
$y_k^{(t)} \in \cup_{l=1}^t \mathbb{Z}^{(l)} \cup \mathbb{Y}$, which is the generalized union of all label sets till the $t$-th session.

Notably, $\mathbb{Z}^{(t_1)} \cap \mathbb{Z}^{(t_2)}=\varnothing$, $\forall t_1, t_2$.  We assume no sample can be retained (unlike rehearsal methods) and the CIL stage only includes (sub-classes of) base classes.
At the $t$-th session, only the support set $\mathbb{S}^{(t)}$ can be used for training.




\begin{figure}[htbp]
    \centering
    \includegraphics[scale=0.5]{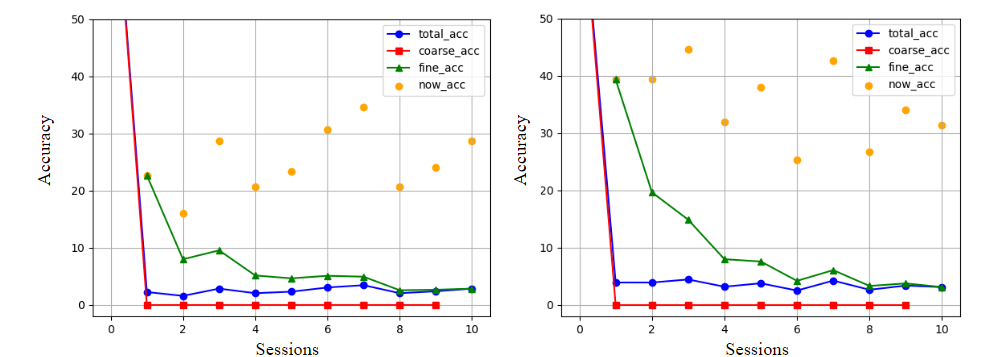}
    \caption{Ablation study of contrastive learning when fine-tuning ResNet12 w/o IL. Left: w/o; right: w/. (CIFAR-100)}
    \label{fig:cgene}
\end{figure}

\begin{figure}[htbp]
    \centering
    \includegraphics[scale=0.5]{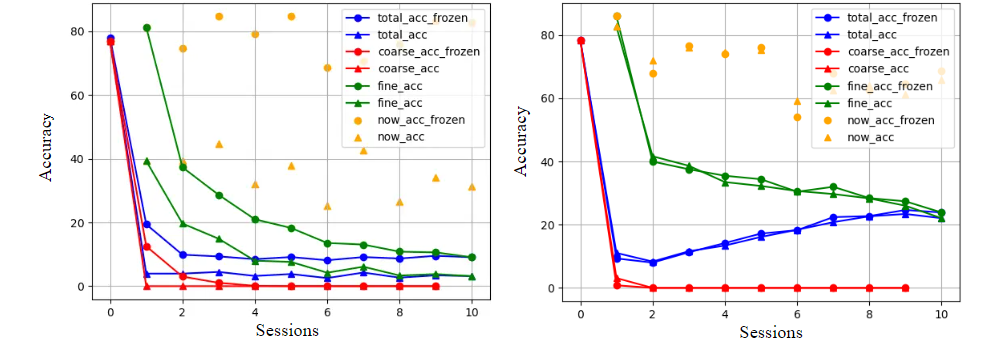}
    \caption{Ablation study of freezing embedding-weights for fine-tuning a contrastive model. Left: when not freezing classifier-weights. Right: when freezing them.  (CIFAR-100)}
    \label{fig:froz}
\end{figure}






\subsection{Embeddings need to be contrastively learned}
\label{sec:ctrneed}
As shown in Fig. \ref{fig:cgene}, straightforward training on coarse labels does not help much the subsequent FSL on fine labels (now\_acc at $\sim 25\%$), while contrastive learning self-supervised by the fine cues does help (now\_acc at $\sim 35\%$). Thus, {\bf coarsely-trained embedding can be generalizable}.

Fig. \ref{fig:froz}-left shows that freezing embedding-weight outperforms not freezing them. It implies the embedding space without any update is generalizable, and that, \emph{if classifier-weights are not frozen, freezing embedding-weights helps}.


\subsection{Freezing weights helps, surely for classifiers}
\label{sec:fzwe}

However, Fig. \ref{fig:froz}-right implies that, \emph{if classifiers weights are frozen, then freezing embedding-weights does not help}.

Comparing left with right of Fig. \ref{fig:froz}, we find that freezing classifier-weights (right) outperforms not doing so (left),
either freezing embedding-weights (circle) or not (triangle). 


\subsection{Weights need to be normalized}

\begin{figure}
    \centering
    \includegraphics[scale=0.23]{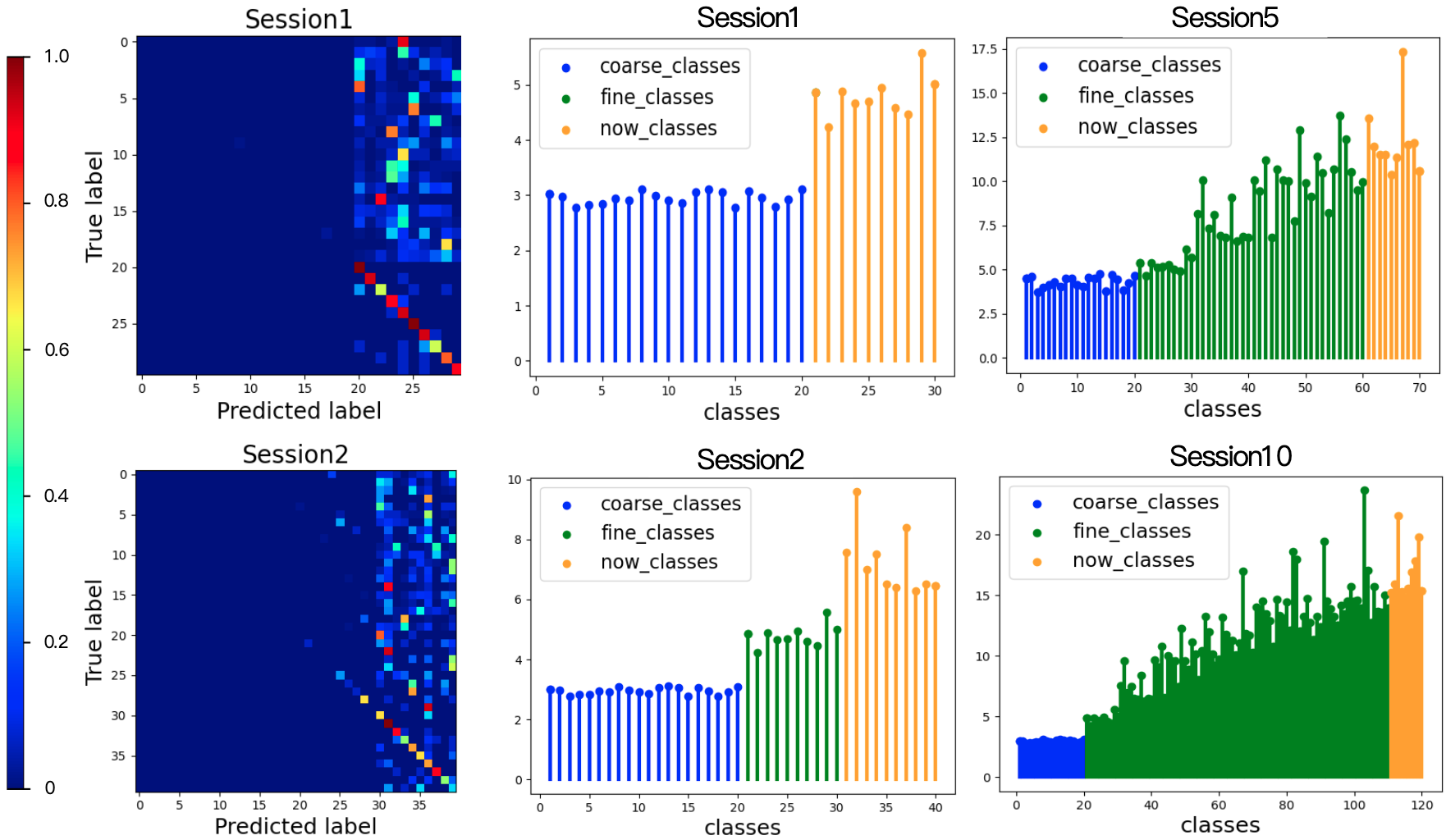}
    \caption{10-way 5-shot confusion matrix (left) and visualization of the norm of raw weights (mid-right) in the last layer for old/new classes. As each session can only access labels of the present classes, a linear classifier will have a larger weight for the current classes' neurons, inducing the queries of previous classes to be likely  assigned into current classes' region (left) in the embedding space. (CIFAR-100) }
    \label{fig:confusion_matrix}
\end{figure}

As shown in Fig. \ref{fig:confusion_matrix}, samples of classes seen in the $1$st session are totally classified to classes seen in the $2$nd session while only samples of the present classes can be correctly classified. 
We plot weight norms to find them grow and propose a conjecture implying a need of normalization. Please see our analysis in the Appendix. 

\noindent {\bf Conjecture 1} (FC weights grow over time). \emph{Let $\|\mathbf{W}^{(t)}\|_F$ denotes the Frobenius norm of the weight matrix formed by all weight vectors in the FC layer for new classes in the $t$-th session. With training converged and norm outliers ignored, it holds that $\|\mathbf{W}^{(t)}\|_F > \|\mathbf{W}^{(t-1)}\|_F, \forall t \in \{1,...,T\}$.}

\subsection{The balance need to be measured}
As already shown in the bottom-right sub-plot of Fig. \ref{fig:diff-sett}, it is  possible to avoid the collapse of coarse-class accuracy, and slow down the accuracy drop of all the previous classes, while still maintaining a high accuracy on present classes.

\begin{figure}[htbp]
\raggedright
\hspace{0.1in}
    \subfloat[living17]{\label {living17}\includegraphics[width=0.5\textwidth]{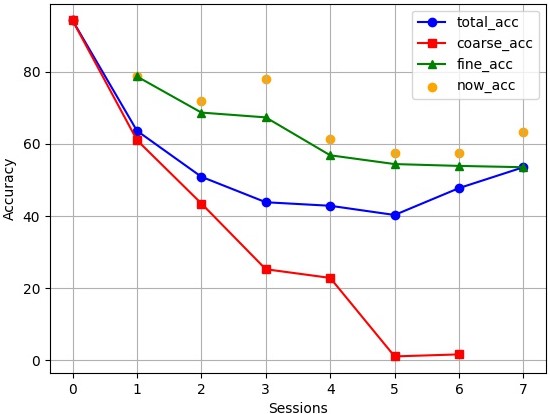}}
	\subfloat[nonliving26]{\label {nonliving26}\includegraphics[width=0.5\textwidth]{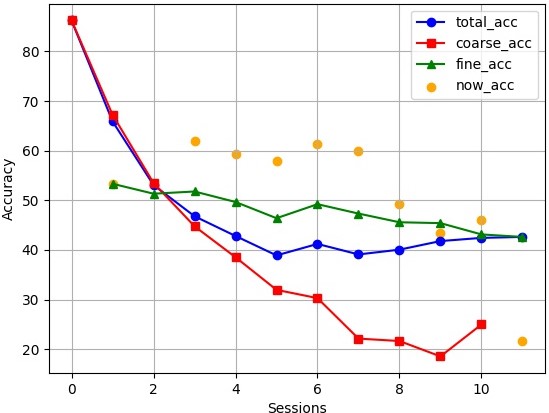}} \\
	\hspace{0.1in}
    \subfloat[entity13]{\label {entity13}\includegraphics[width=0.5\textwidth]{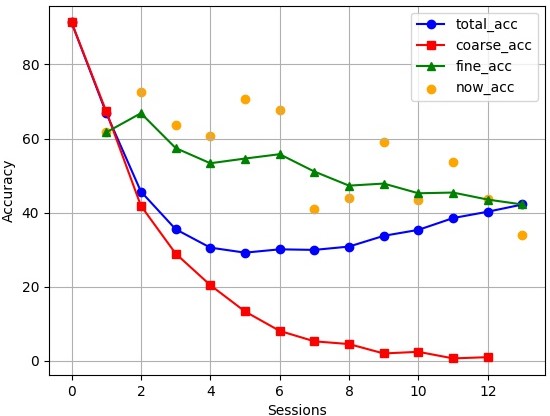}}
	\subfloat[entity30]{\label {entity30}\includegraphics[width=0.5\textwidth]{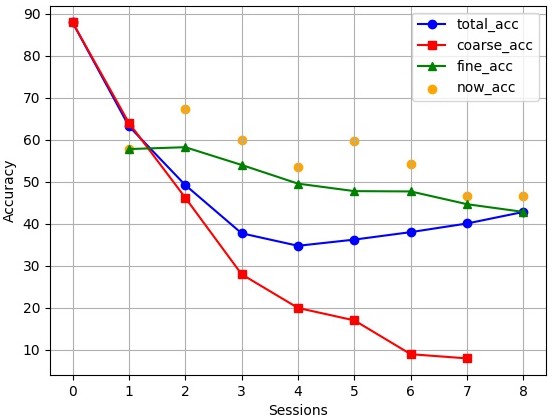}} 
  \caption{Reaching a balance on BREEDS. More in Sec. \ref{sec:exp}.}
	\label{fig:breeds-comp}
\end{figure}

{\bf Old knowledge and current learning can be balanced},
which can be achieved not only on CIFAR-100 but also more generally on BREEDS \cite{santurkar2020breeds}. 
Fig. \ref{fig:breeds-comp} shows the balanced performance on its various subsets. In order to better measure how good the balance is, we also need new overall metrics.

\section{A New Approach: Know-weight (Knowe)}

    
    

\subsection{Learning Embedding-Weights Contrastively}
Now, we elaborate on the Model $0$  of Fig. \ref{fig:proce} about how we train a generalizable base embedding space \cite{rethink20,fshot21}.

We follow ANCOR \cite{cvpr21coarse} to use MoCo \cite{moco} as the backbone, and keep two network streams each of which contains a backbone 
with the last-layer FC replaced by a Multi-Layer Perceptron (MLP). The hidden layer of two  streams' MLP outputs intermediate $\mathbf{q}$ and $\mathbf{k}$, respectively.
Given coarse labels, the total loss is defined as $ \mathcal{L}^c = \mathcal{L}_{Con} + \mathcal{L}_{CE}^c$
where
\begin{equation}
\begin{aligned}
    \mathcal{L}_{Con} = - \sum_{n=1}^N log \frac{exp(\mathbf{q}_n^{\mathrm{T}} \mathbf{k}_n^+ / \tau)}
    {exp(\mathbf{q}_n^{\mathrm{T}} \mathbf{k}_n^+ / \tau) + \sum_{m\neq n} exp(\mathbf{q}_n^{\mathrm{T}} \mathbf{k}_m^- / \tau)},
\end{aligned}
\end{equation}
and $\mathcal{L}_{CE}^c$ is the standard cross-entropy loss that captures the inter-class cues. We also use angular normalization \cite{cvpr21coarse} to improve their synergy.
Note that $m, n$ index samples, $\tau$ is a temperature parameter, $\mathbf{k}_m^{-}$ denotes the intermediate output of the $m$-th sample, a negative sample, in the same class with the $n$-th sample, a positive sample, so as to capture intra-class cues (fine cues), and reduce unnecessary noises to the subsequent fine-grained classification \cite{xu21}. 
$\mathcal{L}_{Con}$ will be small when $\mathbf{q}_n$ is similar with $\mathbf{k}_n^+$ and different from $\mathbf{k}_m^-$. 



\subsection{Normalizing Classifier-Weights}


In the last layer, we set the bias term to $0$. For a sample $\mathbf{x}$, once a neuron has its output logit $o = \mathbf{w}^{\mathrm{T}} \mathbf{f} (\mathbf{x})$ ready, then a Softmax activation function $Smx(\cdot)$ is applied to convert $o$ to a probability so that we can classify $\mathbf{x}$. ($^{\mathrm{T}}$ is transpose)


However, such a inner-product linear classification often favors new classes \cite{hou19}. Instead, we compute the logit using the normalized inner-product \cite{normface} (\emph{a.k.a.}, cosine similarity, cosine normalization \cite{gidaris18,hou19}) as $\tilde{o} = \tilde{\mathbf{w}}^{\mathrm{T}} \tilde{\mathbf{f}} (\mathbf{x})$ where
 $\mathcal{L}_2$-normalized  $\tilde{\mathbf{f}} (\mathbf{x})=\mathbf{f(\mathbf{x})} / \Vert {\mathbf f(\mathbf{x})}\Vert_{2}$ and $\tilde{\mathbf{w}}_i= {\mathbf{w}}_i / \Vert {\mathbf{w}}_i\Vert_{2}$, and then apply Softmax to the rescaled logit $\tilde{o}$ as

\begin{equation}
\begin{aligned}
   p_i(\mathbf{x})=Smx(\tilde{o} / \lambda) =\frac{exp(\tilde{\mathbf{w}}_i^{\mathrm{T}}\tilde{\mathbf{f}} (\mathbf{x}) / \lambda )}
   {\sum_j exp( \tilde{\mathbf{w}}_j^{\mathrm{T}}\tilde{\mathbf{f}} (\mathbf{x}) / \lambda )}
\end{aligned}
\end{equation}
where $i$ is the class index, $\lambda$ is a temperature parameter that rescales the Softmax distribution, as $\tilde{o}$ is ranged of $[-1,1]$.


\subsection{Freezing Memorized Classifier-Weights}

As shown in Fig. \ref{fig:proce}, in the $t$-th incremental session, the task is similar to FSL where a support set $\mathbb S^{(t)}$ is offered to train a model to be evaluated on a query set $\mathbb Q^{(t)}$. However, FSL only evaluates the classification accuracy of the classes appeared in the support set $\mathbb{S}^{(t)}$. In our setting, the query set $\mathbb{Q}^{(t)}$ contains base classes, and all classes in previous support sets. 
As shown in Fig. \ref{fig:froz}, no matter freezing embedding-weights helps or not, it does not hurt. We do so, hoping it to reduce model complexity to avoid over-fitting.

As past samples are not retained, we store the classifier-weights per session to implicitly retain the label information by augmenting a weight matrix $\mathbf{W}$ where in the $t$-th session, we have 
$\mathbf{W}_{[B:E]} = [ \mathbf{w}^{(t)}_1|\mathbf{w}^{(t)}_2 |...|\mathbf{w}_C^{(t)} ]_{d \times C}$
with $B=R+C\cdot(t-1)+1$, $E=R+C\cdot t$ for $t\geq1$, except $\mathbf{W}_{[:R]} = [ \mathbf{w}^{(0)}_1|\mathbf{w}^{(0)}_2 |...|\mathbf{w}_R^{(0)} ]_{d \times R}$ where $d$ is the feature dimension.

 
In the $t$-th session, we minimize the following regularized cross-entropy loss on the support set $\mathbb S^{(t)}$:

\begin{equation}
\begin{aligned}
   \mathcal{L}_{CE}^{\mathbb S^{(t)}} =-\frac{1}{C \cdot K} \sum
   _{n=1} ^{C\cdot K} \sum_{i=1}^{R+t*C} \delta_{y_n^{(t)} = i} log[p_i(\mathbf{x}_n^{(t)})]
\end{aligned}
\end{equation}
where $\delta _{y_n^{(t)}=i}$ is the indicator function and
$p_i(x_n^{(t)})$ is the output probability (\emph{i.e.}, Softmax of logits ) of the $i$-th class.


\subsection{Theoretical Guarantee for Stability-Plasticity}
We extend definitions in \cite{wang21cvpr} 
to set the base of our analysis. Please see also proofs in the Appendix. 

\noindent {\bf Definition 1} (Stability Decay). \emph{For the same input sample, let $\tilde{\mathbf{o}}_i^{(t)}$ denote the output logits of the $i$-th neuron in the last layer in the $t$-th session. After the loss reaches the minimum, we define the degree of stability as $\mathcal{D}=\sum_i(\frac{\tilde{\mathbf{o}}_i^{(T)}-\tilde{\mathbf{o}}_i^{(t)}}{\tilde{\mathbf{o}}_i^{(t)}})^2$}.

\noindent {\bf Definition 2} (Relative Stability). \emph{Given models $\mathbf{\Theta}_a$ and $\mathbf{\Theta}_b$, if $0\le \mathcal{D}_a<\mathcal{D}_b$, then we say $\mathbf{\Theta}_a$ is more stable than $\mathbf{\Theta}_b$.}

Assuming embedding-weights are frozen, then we have:

\noindent {\bf Proposition 1} (Normalizing or freezing weights improves stability; doing both improves the most). \emph{Given $\mathbf{\Theta}_a$, if we only normalize weights of a linear FC classifier, we obtain $\mathbf{\Theta}_b$; if we only freeze them, we obtain $\mathbf{\Theta}_c$; if we do both, we obtain $\mathbf{\Theta}_d$. Then, 
$\mathcal{D}_d < \mathcal{D}_b < \mathcal{D}_a$
and
$\mathcal{D}_d < \mathcal{D}_c < \mathcal{D}_a$.
}


Our second claim is about normalization for plasticity.

\noindent {\bf Proposition 2} (Weights normalized, plasticity remains). \emph{To train our FC classifier, if we denote the loss as $\mathcal{L}(\mathbf{w})$ where $\mathbf{w}$ is normalized, the weight update at each step as $\Delta \mathbf{w}$, and the learning rate as $\alpha$, then we have
$\mathcal{L}(\mathbf{w}-\alpha\Delta\mathbf{w})<\mathcal{L}(\mathbf{w})$.}

Notably, freezing the weights does not affect plasticity.

\subsection{New Overall Performance Measures}
In thi section , we evaluate the model after each session with the query set $\mathbb{Q}^{(t)}$, and report the Top-1 accuracy. The base session only contains coarse labels, and thus is evaluated by the coarse-grained classification accuracy $\mathcal{A}_{c}$.  
We evaluate $\mathcal {A}_{c}$,  the fine-grained accuracy $\mathcal {A}_{f}$, and the total accuracy $\mathcal {A}_{t}$ per incremental session, except the last session when only fine labels are available and $\mathcal {A}_{c}$ is not evaluated. 
We average  $\mathcal A_t$ to obtain an overall performance score as
\begin{equation} 
\label{average acc}
\begin{aligned}
   \bar{\mathcal{A}}=\frac{1}{T+1} \sum _{i=0}^{T} \mathcal{A}^i_{t}.
\end{aligned}
\end{equation}
 
Inspired by \cite{belouadah2020scail}, we define the fine-class forgetting rate
\begin{equation} \label{fine-class drop}
\begin{aligned}
   \mathcal{F}_{f}^t=\frac {\mathcal {A}_{f}^{t-1} -\mathcal {A}_{f}^t}{\mathcal{A}_{f}^{t-1}},
\end{aligned}
\end{equation}
and the forgetting rate for the base coarse class as
\begin{equation} \label{coarse-class drop}
\begin{aligned}
   \mathcal{F}_{c}^t= \frac {\mathcal{A}_{c}^{0} -\mathcal{A}_{c}^t} {\mathcal {A}_{c}^{0}}.
\end{aligned}
\end{equation}


With them, we can evaluate the model with an overall measure to represent the catastrophic forgetting rate as
\begin{equation} \label{total drop}
\begin{aligned}
\mathcal{F}=\frac{1}{T-1} (\sum _{t=2}^{T} \mathcal{F}^t_{f}*\frac{c_t}{N_f}+\sum _{t=1}^{T-1}\mathcal{F}^t_{c}* (1-\frac{c_t}{N_f}))
\end{aligned}
\end{equation}
where $T$ is the number of incremental sessions;   $c_t$ is the number of appeared fine classes until the $t$-th session, and $N_f$ is fine-class total number; $\mathcal A_{c}$ and $\mathcal A_{f}$ are the accuracy of coarse and fine classes 
per session, respectively. 

\section{Experiments}
\label{sec:exp}

\vspace{-1mm}
\subsection{Datasets and Results}
\vspace{-1mm}

 {\bf CIFAR-100} contains $60,000$ 32x32 images from $100$ fine classes, each of which has $500$ training images and 100 test images \cite{cifar}. They can be grouped into $20$ coarse classes, each of which includes $5$ fine classes, \emph{e.g.}, $\emph{trees}$ contains \emph {maple, oak, pine, palm}, and \emph{willow}. The 100 fine classes are  divided into $10$ 10-way 5-shot incremental sessions.
 
 {\bf BREEDS} is derived from ImageNet with class hierarchy re-calibrated by \cite{santurkar2020breeds} and contains $4$ subsets named living17, nonliving26, entity13, and entity30. They have $17$, $26$, $13$, $30$ coarse classes, $4, 4, 20, 8$ fine classes per coarse class, $88$K, $132$K, $334$K, $30$7K training images (224x224), $3.4$K, $5.2$K, $13$K, $12$K test images, respectively. See also Table \ref{tab:data-sett}.
 
{\bf tieredImageNet} (tIN) is a subset of ImageNet and contains $608$ classes \cite{ren18iclr} that are grouped into $34$ high-level super-classes to ensure that the training classes are distinct enough from the test classes semantically. The train/val/test set have $20, 6, 8$ coarse classes, $351, 97, 160$ fine classes, $448$K, $124$K, $206$K images (sized at 84x84), respectively.

\begin{figure}[htbp]
\centering
    \subfloat{\label {total}\includegraphics[width=0.25\textwidth]{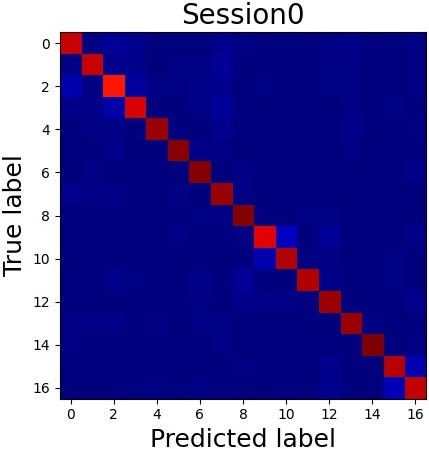}}
	\subfloat{\label {coarse}\includegraphics[width=0.25\textwidth]{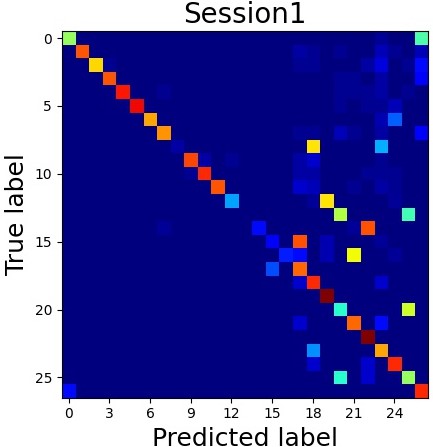}}
    \subfloat{\label {fine}\includegraphics[width=0.25\textwidth]{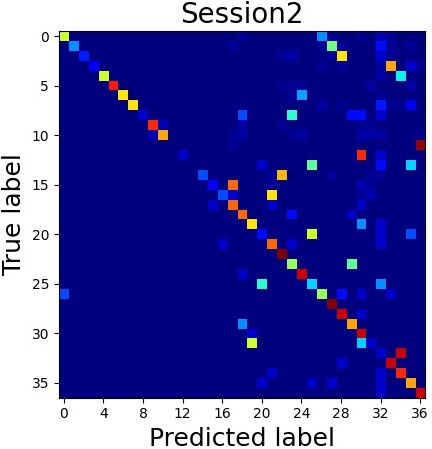}}
    \subfloat{\label {fine}\includegraphics[width=0.25\textwidth]{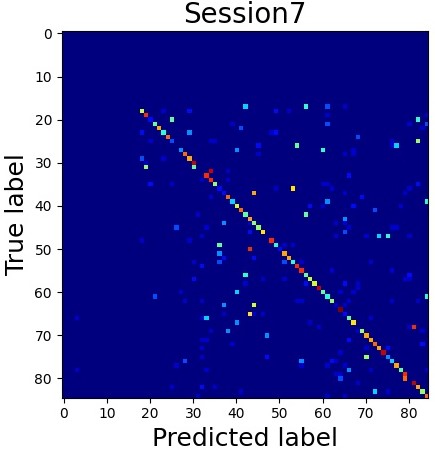}}
  \caption{Confusion matrices of Knowe tested on living17. }
	\label{fig:breeds-confu}
\end{figure}

Table \ref{tab:data-sett} summarizes our performance. Fig. \ref{fig:breeds-comp} shows our separated accuracy on BREEDS, and Fig. \ref{fig:breeds-confu} visualizes confusion matrices to show the evolving of per-class accuracy. 

\begin{table}[hb!]
 \resizebox{\linewidth}{!}{
    \small
    \centering
    \begin{tabular}{ccccccc|cc}
    \toprule
           Dataset & coarse\# & fine\# & total\# & sessions &  way/shot & queries & $\bar{\mathcal{A}}$ & $\mathcal{F}$ \\
    \midrule
         CIFAR-100 & 20 & 100 & 120 & 10 & 10/5 & 15 & 38.50 & 0.42  \\
        living17& 17 & 68 & 85 & 7 & 10/1 & 15 & 54.62 & 0.33   \\
        nonliving26 & 26 & 104 & 130 & 11 & 10/1 & 15 & 48.41 & 0.25  \\
        entity13 & 13 & 260& 273 & 13 & 20/1 & 15 & 41.45 & 0.38  \\
      entity30 & 30 & 240 & 270 & 8 & 30/1 & 15 & 47.79 & 0.32   \\
    tieredImageNet & 20 & 351 & 371 & 10 & 36/5 & 15 & 33.24 & 0.39  \\
    \bottomrule
    \end{tabular}
}
 \centering
\vspace{2mm}
 \caption{Dataset setting and performance. \# is class num.}
    \label{tab:data-sett}
\end{table}


\subsection{Implementation Details} 

We use ResNet-50 on BREEDS, ‘-12’ on CIFAR-100 and ‘-12’ on tIN, train $\mathbf{\Theta}^{(0)}$ except FC using ANCOR, use SGD with a momentum $0.9$, as well as set weight decay to 5e-4, batch size to $256$, $\tau$ to $0.2$, and $\lambda$ to $0.5$. The learning rate is $0.12$ for $\mathbf{\Theta}^{(0)}$, and is $0.1$ for $\mathbf{\Theta}^{(1)}$,$\mathbf{\Theta}^{(2)}$, \emph{etc.} for $200$ epochs.

\subsection{Ablation Study} 

\begin{figure}[htbp]
\centering
    \subfloat[Base contrastive learning?]{\label {fig:embedding comparation}\includegraphics[width=0.5\textwidth]{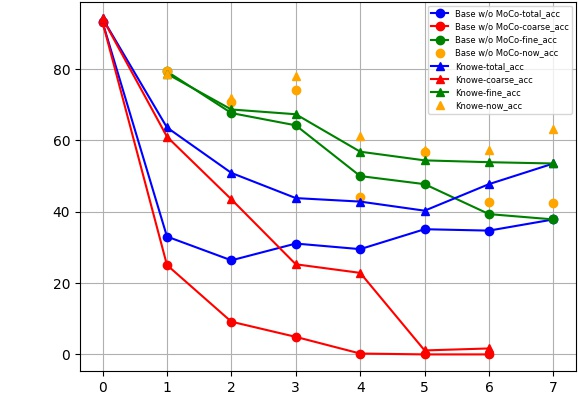}}
	\subfloat[Freezing embedding-weights?]{\label {fig:frozen embedding}\includegraphics[width=0.5\textwidth]{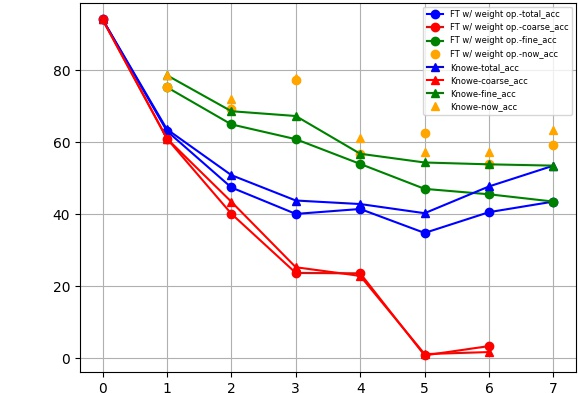}}\\
    \subfloat[Normalizing classifier-weights?]{\label {fig:cosine normalization}\includegraphics[width=0.5\textwidth]{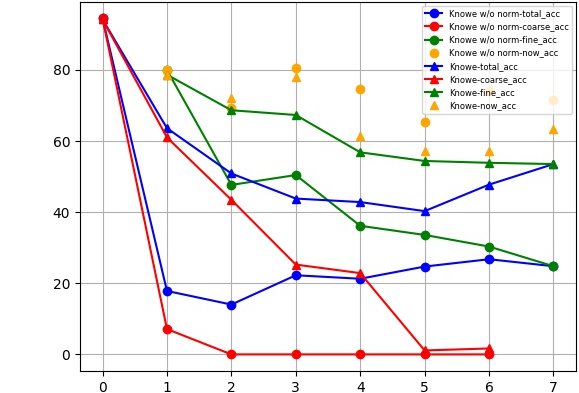}}
    \subfloat[Freezing classifier-weights?]{\label {fig:FT fc}\includegraphics[width=0.5\textwidth]{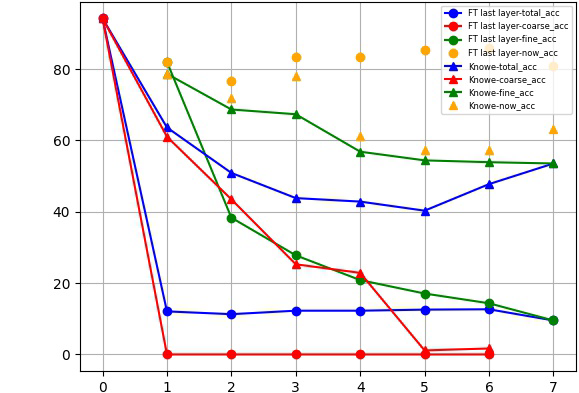}}
    \caption{$4$-factor ablation study on living17, separated \emph{acc}. }
	\label{fig:breeds-cross-comp}
\end{figure}

\textbf{Impact of base contrastive learning.}  
Fig. \ref{fig:cgene} already illustrates its benefit for a simple model without any weight operation. 
As shown in Fig \ref{fig:embedding comparation}, Knowe also obtains a better performance than not using MoCo in Knowe's base, which verifies that the contrastively-learned base model helps fine-grained recognition. Starting from almost the same fine accuracy in the $2$nd session, the gap between w/ MoCo and w/o MoCo increases, as the former stably outperforms the latter on current classes.
It verifies that the former can learn more fine knowledge than the latter. Given that there are only a few fine-class samples, the extra fine-grained knowledge is likely from the contrastively-learned base model.


\textbf{Impact of freezing embedding-learner weight} (decoupling). 
It has been verified in Sec.\ref{sec:ctrneed} that, \emph{if neither freezing nor normalizing classifier-weights, freezing the embedding-weights helps}. We have a {\bf Conjecture 2}: \emph{$\neg p \land \neg q \Rightarrow r$ where $p$ is a premise that classifier-weights are normalized, $q$ is another that classifier-weights are frozen, and $r$ is a conclusion that freezing embedding-weights improves Knowe's performance}.  
However, Fig. \ref{fig:frozen embedding} illustrates that, for Knowe, freezing embedding-weights induces a slightly better performance than not freezing them.
It implies, \emph{if classifier-weights are normalized and frozen, then freezing the embedding-weights does not help} ($p \land q \Rightarrow \neg r$), which is shown by small changes of ${\bar{\mathcal{A}}}$ and $\mathcal{F}$ in Table \ref{tab:perf-comp}.

\textbf{Impact of normalizing classifier-weight}. 
Fig \ref{fig:confusion_matrix} has already shown that, with a linear classifier, the weight norms of new classes totally surpass the weight norms of previous classes, which causes that the linear classifier biases towards new classes (\emph{i.e.}, any sample of previous class can be classified as a new class). That implies a need of normlizing the classifer-weights.
As shown in Fig \ref{fig:cosine normalization}, when we freeze weights of previous classes and only tune the weights of new classes without normalization, the model performs stably worse than Knowe with normalization,
which verifies that normalizing classifier-weights plays a positive role.

\textbf{Impact of freezing memorized classifier-weights}. As shown in Fig \ref{fig:FT fc}, there is severe CF of both fine and coarse knowledge when not freezing the weights of previously-seen classes, which implies that little knowledge is retained.
Although embedding-weights are frozen and classifier-weights are normalized, the coarse knowledge is totally forgotten. It implies that, \emph{if classifier-weights are normalized and yet not frozen, freezing the embedding-weights does not help} ($p \land \neg q \Rightarrow \neg r$).
It can be explained that fine-tuning on a few samples normally induces little change to the embedding-weights and yet great change to classifier-weights.
Moreover, the model without freezing classifier-weights performs much worse than Knowe that freezes previous weights. The gap of the fine accuracy increases over time and is larger than the gap of the present accuracy. It implies that they also differ in the performance of previous fine classes, which is the CF of learned fine knowledge.

{\bf More about freezing embedding-weights}. In Sec.\ref{sec:fzwe}, we know $\neg p \land q \Rightarrow \neg r$.  Thus, we have a {\bf Conjecture 3}: \emph{$p \vee q \Leftrightarrow \neg r$}, meaning \emph{if and only if classifier-weights are either normalized or frozen, then freezing embedding-weights does not help}. Please see also the Appendix for the analysis.


{\bf Overall finding}. A decent \emph{now\_acc} seems to be a condition for weight freezing and normalization to be effective.

\begin{table*}[h!]
\resizebox{\linewidth}{!}{
\begin{tabular}{lccccccccccccc|cc}
\toprule
\multicolumn{1}{c}{\multirow{2}{*}{Mehtod}} & \multicolumn{1}{c}{\multirow{2}{*}{Contr. learn.}} & \multicolumn{1}{c}{\multirow{2}{*}{Decoupled}}& \multicolumn{1}{c}{\multirow{2}{*}{Normalization}} &  \multicolumn{1}{c}{\multirow{2}{*}{Frozen}} &&\multicolumn{8}{c}{\multirow{1}{*}{Total accuracy per session}} &  \multicolumn{1}{|c}{\multirow{2}{*}{$\bar{\mathcal A}\uparrow$}} & \multicolumn{1}{c}{\multirow{2}{*}{$\mathcal{F}\downarrow$ }} \\
& &\multicolumn{1}{c}{}  &\multicolumn{1}{c}{}  &\multicolumn{1}{c}{} &0 &1 &2 &3 &4 &5 &6 &7 &8 &\multicolumn{1}{c}{} &\multicolumn{1}{c}{} \\
\midrule
(a) Base w/o MoCo  &  & \checkmark & \checkmark &\checkmark & 93.18 & 33.04 & 26.37 & 31.08 & 29.51 & 35.10 & 34.71 & 37.84 & N/A  &40.10&0.50\\
(b) FT w/ weight op.  & \checkmark &  &  \checkmark   &\checkmark & 94.21 &63.14 &47.45 &40.10 &41.47 &34.80 & 40.59 & 43.53 &N/A &50.66&0.35\\
(c) Knowe w/o norm.  & \checkmark & \checkmark &   & \checkmark  & 94.50 & 17.84 &14.02 &22.26  & 21.28 & 24.71 & 26.77 &24.80 &N/A &30.77&0.57\\
(d) FT last layer   & \checkmark & \checkmark & \checkmark  &   & 94.21 & 12.06 &11.28 &12.26  &12.26 &12.55 &12.65 &9.51 &N/A &22.09&0.66\\
\midrule
LwF+ \cite{li2016learning}    & \checkmark &  &  & &\bf 94.50 & \emph{61.47} & \emph{44.61} & 27.45 & 19.12 &11.28  &6.37  &4.22  &N/A &33.63&0.51\\
ScaIL \cite{belouadah2020scail}    & \checkmark &  &    & &\bf 94.50 & 38.63 &25.59 &31.08 &30.29 &35.10 &37.84 &41.08 &N/A&41.76&0.48\\
Weight Align+ \cite{zhao20cvpr}& \checkmark & \checkmark &  & \checkmark & \bf94.50 & 50.98 & 37.94 & \emph{38.43} & \emph{37.06} & \emph{35.20} & \emph{39.80} & \emph{43.24} &N/A & \emph{47.14}& \emph{0.40} \\
Subsp. Reg.+ \cite{akyurek2021subspace}& \checkmark & \checkmark &   & \checkmark &\bf 94.50 & 59.41 & 39.51 & 33.43 & 29.31 & 25.59 & 27.84 & 26.47 &N/A &42.01&\emph{0.40}\\
Knowe (Ours)   & \checkmark & \checkmark &  \checkmark &  \checkmark & \emph{94.21} & \bf63.63 & \bf 50.88 &{\bf 43.82} &{\bf 42.84} & \bf 40.29 & \bf47.75 & \bf53.53 &N/A &{\bf 54.62}& \bf 0.33\\
\midrule
ANCOR \cite{cvpr21coarse}&\checkmark&&&&94.50&11.86&11.18&12.35&11.77&12.55&10.78&9.02&N/A&21.75&0.66 \\
Jt. train. (upp. bd.)    & \checkmark & \checkmark &   \checkmark & \checkmark & 94.21 & 63.63 & 58.53 & 52.26 & 46.28 & 47.75 & 36.96 & 42.75 &N/A & 55.29& 0.25\\
\bottomrule
\midrule
LwF+\cite{li2016learning}    & \checkmark &  &  & &\bf89.48&\bf65.03&\emph{48.69}&22.72&9.36&6.03&4.61&2.86&3.33  &28.01&0.47\\
ScaIL\cite{belouadah2020scail}    & \checkmark &  &    & &\bf89.48&39.25&25.50&22.44&23.69&25.75&30.81&32.08&35.25   &36.03&0.48\\
Weight Align+ \cite{zhao20cvpr}& \checkmark & \checkmark &  & \checkmark &\bf89.48&47.36&37.06&\emph{31.72}&\emph{30.56}&\emph{32.28}&\emph{34.11}&\emph{36.39}&\emph{37.06}  &\emph{41.78}&\emph{0.42}\\
Subsp. Reg.+  \cite{akyurek2021subspace}& \checkmark & \checkmark &   & \checkmark &\bf89.48& 42.39& 28.94& 20.86& 16.14& 16.44& 16.75& 16.17& 16.06&29.25&0.48\\
Knowe (Ours)   & \checkmark & \checkmark &  \checkmark &  \checkmark &\emph{87.90}&\emph{63.22}&\bf49.22&\bf37.75&\bf34.78&\bf36.25&\bf38.03&\bf40.08&\bf42.83  &\bf47.79&\bf0.32\\
\midrule
ANCOR\cite{cvpr21coarse} &\checkmark &&&&89.48&8.67&8.28&9.50&6.83&8.75&9.53&8.19&8.69&17.55&0.61\\
Jt. train. (upp. bd.)    & \checkmark & \checkmark &   \checkmark & \checkmark &87.90&63.22&56.56&53.72&47.36&44.78&41.61&38.06&36.75   &52.22&0.20\\
\bottomrule
\end{tabular}}
\caption{Ablation study of 4 factors and comparison with others on BREEDS living17 (top) and entity30 (bottom). Best seen on computer.}
\label{tab:perf-comp}
\end{table*}

\begin{table*}[h!]
\resizebox{\linewidth}{!}{
\begin{tabular}{lcccccccccccccccccc|cc}
\toprule
\multicolumn{1}{c}{Method} & \multicolumn{1}{c}{Contr. learn.} & \multicolumn{1}{c}{Decoupled}& \multicolumn{1}{c}{Normalization} &  \multicolumn{1}{c}{Frozen} &0&1&2&3&4&5&6&7&8&9&10&11&12&13 & \multicolumn{1}{c}{$\bar{\mathcal A}\uparrow$} & \multicolumn{1}{c}{$\mathcal{F}\downarrow$} \\
\midrule
LwF+ \cite{li2016learning}    & \checkmark &  &  & & \bf 86.94&\emph{65.51}& \bf 58.14&\emph{44.17}&22.76&14.36&9.68&6.92&5.90&5.19&5.32&3.40 &N/A&N/A &27.36&0.38\\
ScaIL \cite{belouadah2020scail}    & \checkmark &  &    & & \bf 86.94&36.09&24.10&21.47&23.27&23.65&27.95&31.80&34.23&36.09&\emph{37.76}&\emph{38.14} &N/A&N/A &35.12&0.43\\
Weight Align.+ \cite{zhao20cvpr}& \checkmark & \checkmark &  & \checkmark &\bf 86.94&61.41&46.03&40.00&35.77&\emph{34.10}&\emph{35.96}&\emph{33.21}&\emph{35.51}&\emph{36.60}&37.56&37.76 &N/A&N/A  &\emph{43.40}&\emph{0.29}\\
Subsp. Reg.+ \cite{akyurek2021subspace}& \checkmark & \checkmark &   & \checkmark &\bf 86.94 &63.59&52.56&42.95&\emph{35.96}&31.41&28.01&26.15&23.27&19.68&19.36&20.19 &N/A&N/A &37.51&\bf 0.25\\
Knowe (Ours)   & \checkmark & \checkmark &  \checkmark &  \checkmark &\emph{86.23}&\bf65.90&\emph{53.08}&\bf 46.80 &\bf 42.82&\bf 38.91&\bf 41.22&\bf 39.10&\bf 40.06&\bf41.80&\bf42.44&\bf42.63  &N/A&N/A&\bf 48.41&\bf 0.25\\
\midrule
ANCOR \cite{cvpr21coarse} &\checkmark &&&& \bf 86.94 &5.83&6.03&6.92&5.90&6.60&7.63&7.05&7.05&7.50&7.44&2.63&N/A&N/A&13.13 & 0.61\\
Jt. train. (upp. bd.)    & \checkmark & \checkmark &   \checkmark & \checkmark & 86.23& 65.90 & 60.51 & 59.04 & 53.53 & 53.85 & 46.73 & 46.60 & 43.85 & 36.67 & 37.31 & 36.80 &N/A&N/A& 52.25& 0.16\\
\bottomrule
\midrule
LwF+\cite{li2016learning}    & \checkmark &  &  & &\bf92.03 & \emph{59.10} &\emph{43.64} &18.49 &10.49 &6.82  &3.59  &2.54  &3.10 &2.56 &2.10 &2.23 &1.77 &1.54  &17.86&0.52\\
ScaIL\cite{belouadah2020scail}    & \checkmark &  &    & &\bf92.03 &37.10 &13.92 &13.36 &14.87 &18.36 &21.72 &23.28 &24.33 &27.62 &29.59 &31.54 &32.36 & 34.08  &29.58&0.49\\
Weight Align+ \cite{zhao20cvpr}& \checkmark & \checkmark &  & \checkmark &\bf92.03 &36.74 &24.15 &\emph{20.51} &\emph{22.31} &\emph{24.82} &\emph{26.41} &\emph{26.85} &\emph{27.26} &\emph{31.49} &\emph{32.26} &\emph{35.28} &\emph{36.72} &\emph{37.69}  &\emph{33.89}&0.46\\
Subsp. Reg.+ \cite{akyurek2021subspace}& \checkmark & \checkmark &   & \checkmark &\bf92.03& 52.72& 28.95& 15.92& 12.08& 10.82& 10.90& 11.49& 12.05& 12.03& 11.77& 11.72& 12.54&14.36 &22.10&\emph{0.45}\\
Knowe (Ours)   & \checkmark & \checkmark &  \checkmark &  \checkmark & \emph{91.35} &{\bf 66.90} &{\bf 45.69} &{\bf 35.54} &{\bf 30.56} &{\bf 29.21} &\bf30.10 &\bf29.95 &\bf30.85 &\bf33.74 &\bf35.36 &\bf38.54 &\bf40.26 &\bf42.21  &\bf41.45&\bf0.38\\
\midrule
ANCOR\cite{cvpr21coarse} &\checkmark &&&&92.03&5.36&5.67&5.49&5.18&6.51&5.82&4.80&5.39&6.28&5.36&5.13&5.26&5.62&11.71&0.57\\
Jt. train. (upp. bd.)    & \checkmark & \checkmark &   \checkmark & \checkmark &91.35& 66.90& 57.54& 49.92& 50.59& 48.64& 47.69& 44.41& 41.72& 39.13& 39.62& 40.72& 38.49& 37.26 &49.57&0.24\\
\bottomrule
\end{tabular}}
\caption{Comparison with others on BREEDS nonliving26 (top) and entity13 (bottom). {\bf Bold} is the best, \emph{slanted} is $2$nd. Best seen on computer.}
\label{tab:perf-comp-nonliving26}
\end{table*}

\begin{table*}[h!]
\resizebox{\linewidth}{!}{
\begin{tabular}{lccccccccccccccc|cc}
\toprule
\multicolumn{1}{c}{Method} & \multicolumn{1}{c}{Contr. learn.} & \multicolumn{1}{c}{Decoupled}& \multicolumn{1}{c}{Normalization} &  \multicolumn{1}{c}{Frozen} &0&1&2&3&4&5&6&7&8&9&10 & \multicolumn{1}{c}{$\bar{\mathcal A}\uparrow$} & \multicolumn{1}{c}{$\mathcal{F}\downarrow$ } \\
\midrule
LwF+\cite{li2016learning}    & \checkmark &  &  & &\bf 78.39 	&\bf 41.87 	&28.00	&23.80	&14.93	&10.53	&8.00	&8.80	&6.47	&7.33	&6.73&21.35&0.51\\
ScaIL\cite{belouadah2020scail}    & \checkmark &  &    & &\bf 78.39	&14.47	&14.13	&18.07	&21.00	&25.20	&26.20	&31.87	&\emph{32.60}	&36.53	&38.20&30.61&0.52\\
Weight Align+ \cite{zhao20cvpr}& \checkmark & \checkmark &  & \checkmark &\bf 78.39   	&13.20	&14.13	&18.20	&21.20	&24.60	&\emph{26.93}	&\emph{32.33}	&\emph{32.60}	&\emph{38.93}	&\emph{38.46}&30.82&0.53\\
Subsp. Reg.+ \cite{akyurek2021subspace}& \checkmark & \checkmark &   & \checkmark &\bf 78.39	&\emph{41.47}	&\bf 31.80	&\bf 32.87	&\emph{26.73}	&\emph{25.73}	&25.27	&26.73	&24.27	&25.73	&24.00&\emph{33.00}&\emph{0.43} \\
Knowe (Ours)   & \checkmark & \checkmark &  \checkmark &  \checkmark &\emph{72.07}	&36.00	&\emph{28.13}	&\emph{30.27}	&\bf 32.20	&\bf 31.20	&\bf 30.93	&\bf 36.33	&\bf 39.27	&\bf 43.20	&\bf 43.93&\bf 38.50&\bf 0.42\\
\midrule
ANCOR\cite{cvpr21coarse} &\checkmark &&&& 78.39	&7.93	&7.13	&8.27	&7.80	&8.60	&6.40	&7.53	&6.93	&8.20	&8.33&14.14&0.59\\
Jt. train. (upp. bd.)    & \checkmark & \checkmark &   \checkmark & \checkmark &72.07	&36.00	&37.07	&40.27	&40.13	&41.33	&38.60	&41.13	&40.47	&41.40	&43.47&42.90&0.33\\
\bottomrule
\midrule
LwF+\cite{li2016learning}    & \checkmark &  &  & &\bf87.64& \bf69.36& 13.88& 4.22& 4.05& 4.03& 3.02& 2.74& 1.44& 1.05& 1.06&17.50&0.55\\
ScaIL\cite{belouadah2020scail}    & \checkmark &  &    & &\bf87.64& 48.51&\bf33.12&\bf26.15&\bf22.66&\emph{22.77}& 23.42& 22.72& 23.38& 25.17& 26.65&\emph{32.93}&\emph{0.40}\\
Weight Align+ \cite{zhao20cvpr}& \checkmark & \checkmark &  & \checkmark &\bf87.64& 25.13& 18.63& 18.37& 20.08& 22.20& \emph{24.22}&\bf24.73&\emph{26.71}&\emph{29.00}&\emph{30.45}&29.74&0.48\\
Subsp. Reg.+ \cite{akyurek2021subspace}& \checkmark & \checkmark &   & \checkmark &\bf87.64&\emph{49.73}&\emph{32.06}& 24.35& 20.95& 20.76& 20.84& 21.12& 21.79& 23.15& 24.31&31.52&0.42\\
Knowe (Ours)   & \checkmark & \checkmark &  \checkmark &  \checkmark &\emph{76.15}&48.24&30.60&\emph{25.60}&\emph{22.34}&\bf23.48&\bf24.79&\emph{24.69}&\bf27.65&\bf30.26&\bf31.87&\bf33.24 &\bf0.39\\
\midrule
ANCOR\cite{cvpr21coarse} &\checkmark &&&&87.64& 7.10& 6.69& 6.55& 6.36& 6.57& 6.42& 6.55& 6.55& 6.40& 5.17&13.82&0.61\\
Jt. train. (upp. bd.)   & \checkmark & \checkmark &   \checkmark & \checkmark &76.15& 48.24& 39.89& 34.09& 32.21& 30.85& 28.81& 29.86& 28.57& 28.74& 29.06&36.95&0.32\\
\bottomrule
\end{tabular}}
\vspace{1mm}
\caption{Comparison with others on CIFAR-100 (top table) and tieredImageNet (bottom table). '+' means improvement. Best seen on computer.}
\label{tab:perf-comp-tiered-4}
\end{table*}

\begin{figure*}[h!]
    \centering
    \includegraphics[scale=0.4]{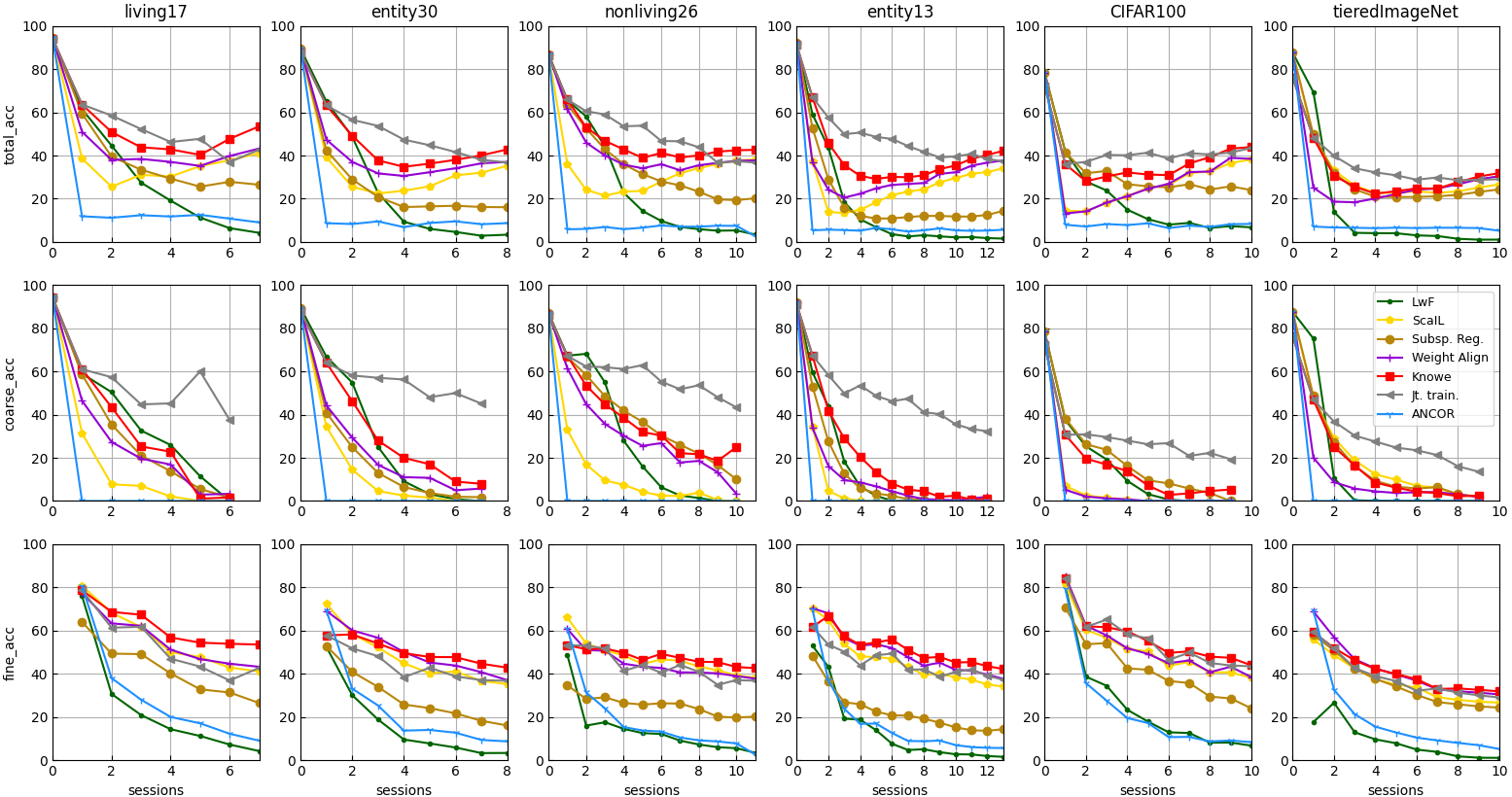}
    \caption{Separated accuracy comparison on all datasets. Top-down: total, corese, fine; red is Knowe, grey is joint training.}
    \label{fig:comparison on four datasets}
\end{figure*}

\subsection{Performance Comparison and Analysis}

Table \ref{tab:perf-comp},\ref{tab:perf-comp-nonliving26},\ref{tab:perf-comp-tiered-4} and Fig. \ref{fig:comparison on four datasets} compare Knowe with SOTA FSCIL/IL methods including LwF \cite{li2016learning}, ScaIL \cite{belouadah2020scail}, Weight Aligning \cite{zhao20cvpr} and Subspace Regularizers (Sub. Reg.) \cite{akyurek2021subspace}. Joint training is non-IL and an \emph{acc} upper bound in principle. 

{\bf Overall metrics}: average \emph{acc} $\bar {\mathcal A}$ and forgetting rate $\mathcal{F}$. As presented in Table \ref{tab:perf-comp}, \ref{tab:perf-comp-nonliving26}, \ref{tab:perf-comp-tiered-4}, Knowe has the smallest $\mathcal{F}$ and the largest $\bar {\mathcal{A}}$ on all datasets. From both metrics,
Weight Aligning ranks $2$nd on BREEDS, Sub. Reg. ranks $2$nd on CIFAR-100, and ScaIL ranks $2$nd on tIN.
There is a consistency of two metrics.
LwF often has poor numbers, which implies that, with no samples retained, KD does not help. 

{\bf Total accuracy} per session decreases over time yet slower and slower for Knowe and SOTA methods. However, outstanding ones decrease first and then rise, because that the proportion of fine classes  in the query set gets higher and their accuracy plays a leading role in the total accuracy. Knowe is the best, with a strong rising trend, which satisfies the aim of CIL the most and envisions Knowe continuing performing well when more sessions are added (Table \ref{tab:perf-comp-nonliving26}) . Sub. Reg. and Weight Align. often have $2$nd-best numbers (both freeze weights); ScaIL and LwF occasionally do.

{\bf Coarse class accuracy} decreases over time unavoidably (see Fig. \ref{fig:comparison on four datasets}), while Knowe and SOTA methods slow down the decay, with comparable rates. As IL methods, Weight Aligning, ScaIL, and LwF do not forget  knowledge totally although they do not operate weights as done by Knowe. As an non-IL approach, ANCOR totally forgets old knowledge from the $1$st session because it fine-tunes on the few fine shots without any extra operation to retain coarse knowledge. 
The joint training on all fine classes till the present is non-IL, and in principle should bound the fine-class performance. Interestingly, it also suffers less from coarse \emph{acc} decay, the rate of which is much lower (Fig. \ref{fig:comparison on four datasets}). Differently, the cause can be imbalance between increasing fine classes and existing coarse classes. 
Knowe's performance is very competitive and indeed bounded by joint training.

{\bf Fine class's total accuracy} normally decreases over time yet slower and slower for Knowe and SOTA methods (Fig. \ref{fig:comparison on four datasets}), and can be maintained in a similar range for most methods, among which Knowe often stays the highest, ScaIl and Weight Aligning are in the middle, Sub Reg. often stays in a low level, and LwF and ANCOR perform stably the worst. 
Knowe is the most balanced, while Sub. Reg. biases towards stability that is its drawback. Joint training does not necessarily bound the accuracy, possibly due to few shots.

\section{Conclusion}
In this paper, we present a challenging new problem, new metrics, insights, and an approach that solves it well in the sense of getting more balanced performance than the state-of-the-art approaches.

While it is not new to freeze or normalize weights, we are unaware of them previously being presented as a principled approach (to CIL) that is as simple as fine-tuning. It makes pre-trained big models more useful for finer-grained tasks. 

For C2FSCIL with a linear classifier, {\bf weights seem to be the knowledge}. However, how generic are our findings in practice? Can they be applied to general FSCIL? If yes, we are more comfortable with that answer, but then how does a class hierarchy make a difference?
Future work will include examining those questions, non-linear classifiers, and so on.


{\small
\bibliographystyle{unsrt}
\bibliography{egbib}

\begin{thebibliography}{10}

\bibitem{mai21survey}
Zheda Mai, Ruiwen Li, Jihwan Jeong, David Quispe, Hyunwoo Kim, and Scott
  Sanner.
\newblock Online continual learning in image classification: An empirical
  survey.
\newblock {\em ArXiv preprint:2101.10423}, 2021.

\bibitem{cifar}
Alex Krizhevsky.
\newblock Learning multiple layers of features from tiny images.
\newblock {\em Unvieristy of Toronto: Technical Report}, 2009.

\bibitem{cog99}
Robert M.French.
\newblock Catastrophic forgetting in connectionist networks.
\newblock {\em Trends in Cognitive Sciences}, 3, 1999.

\bibitem{kirkpatrick2017overcoming}
James Kirkpatrick, Razvan Pascanu, Neil Rabinowitz, Joel Veness, Guillaume
  Desjardins, Andrei~A Rusu, Kieran Milan, John Quan, Tiago Ramalho, Agnieszka
  Grabska-Barwinska, et~al.
\newblock Overcoming catastrophic forgetting in neural networks.
\newblock {\em Proceedings of the National Academy of Sciences},
  114(13):3521--3526, 2017.

\bibitem{colt21coarse}
Dimitris Fotakis, Alkis Kalavasis, Vasilis Kontonis, and Christos Tzamos.
\newblock Efficient algorithms for learning from coarse labels.
\newblock In {\em 34th Annual Conference on Learning Theory}, 2021.

\bibitem{cha21}
Hyuntak Cha, Jaeho Lee, and Jinwoo Shin.
\newblock Co2l: Contrastive continual learning.
\newblock In {\em ICCV}, 2021.

\bibitem{mermillod2013stability}
Martial Mermillod, Aur{\'e}lia Bugaiska, and Patrick Bonin.
\newblock The stability-plasticity dilemma: Investigating the continuum from
  catastrophic forgetting to age-limited learning effects.
\newblock {\em Frontiers in Psychology}, 4, 2013.

\bibitem{wang21cvpr}
Shipeng Wang, Xiaorong Li, Jian Sun, and Zongben Xu.
\newblock Training networks in null space of feature covariance for continual
  learning.
\newblock In {\em CVPR}, 2021.

\bibitem{cog89}
M~McCloskey and N~Cohen.
\newblock Catastrophic interference in connectionist networks: the sequential
  learning problem.
\newblock {\em The Psychology of Learning and Motivation}, 24:109--164, 1989.

\bibitem{cvpr21coarse}
Guy Bukchin, Eli Schwartz, Kate Saenko, Ori Shahar, Rogerio Feris, Raja Giryes,
  and Leonid Karlinsky.
\newblock Fine-grained angular contrastive learning with coarse labels.
\newblock In {\em CVPR}, 2021.

\bibitem{xu21}
Yuanhong Xu, Qi~Qian, Hao Li, Rong Jin, and Juhua Hu.
\newblock Weakly supervised representation learning with coarse labels.
\newblock In {\em ICCV}, 2021.

\bibitem{yang21}
Jinhai Yang, Hua Yang, and Lin Chen.
\newblock Towards cross-granularity few-shot learning: Coarse-to-fine
  pseudo-labeling with visual-semantic meta-embedding.
\newblock In {\em ACM Conference on Multimedia}, 2021.

\bibitem{riste15}
Marko Ristin, Juergen Gall, Matthieu Guillaumin, and Luc~Van Gool.
\newblock From categories to subcategories: Large-scale image classification
  with partial class label refinement.
\newblock In {\em CVPR}, 2015.

\bibitem{riste14}
Marko Ristin, Matthieu Guillaumin, Juergen Gall, and Luc~Van Gool.
\newblock {Incremental Learning of NCM Forests for Large-Scale Image
  Classification}.
\newblock In {\em CVPR}, 2014.

\bibitem{open15}
Abhijit Bendale and Terrance Boult.
\newblock Towards open world recognition.
\newblock In {\em CVPR}, 2015.

\bibitem{iCaRL}
Sylvestre-Alvise Rebuff, Alexander Kolesnikov, Georg Sperl, and Christoph~H.
  Lampert.
\newblock {iCaRL: Incremental Classifier and Representation Learning}.
\newblock In {\em CVPR}, 2017.

\bibitem{il-svm}
Ilja Kuzborskij, Francesco Orabona, and Barbara Caputo.
\newblock From n to n+1: Multiclass transfer incremental learning.
\newblock In {\em CVPR}, 2013.

\bibitem{de21survey}
Matthias Delange, Rahaf Aljundi, Marc Masana, Sarah Parisot, Xu~Jia, Ales
  Leonardis, Greg Slabaugh, and Tinne Tuytelaars.
\newblock A continual learning survey: Defying forgetting in classification
  tasks.
\newblock {\em IEEE Transactions on Pattern Analysis and Machine Intelligence},
  2021.

\bibitem{Tao_2020_CVPR}
Xiaoyu Tao, Xiaopeng Hong, Xinyuan Chang, Songlin Dong, Xing Wei, and Yihong
  Gong.
\newblock Few-shot class-incremental learning.
\newblock In {\em CVPR}, 2020.

\bibitem{tao20eccv}
Xiaoyu Tao, Xinyuan Chang, Xiaopeng Hong, Xing Wei, and Yihong Gong.
\newblock Topology-preserving class-incremental learning.
\newblock In {\em ECCV}, 2020.

\bibitem{rethink20}
Yonglong Tian, Yue Wang, Dilip Krishnan, Joshua~B. Tenenbaum, and Phillip
  Isola.
\newblock Rethinking few-shot image classification: a good embedding is all you
  need?
\newblock In {\em ECCV}, 2020.

\bibitem{gidaris18}
Spyros Gidaris and Nikos Komodakis.
\newblock Dynamic few-shot visual learning without forgetting.
\newblock In {\em CVPR}, 2018.

\bibitem{ren19}
Mengye Ren, Renjie Liao, Ethan Fetaya, and Richard~S. Zemel.
\newblock Incremental few-shot learning with attention attractor networks.
\newblock In {\em NeurIPS}, 2019.

\bibitem{dong21}
Songlin Dong, Xiaopeng Hong, Xiaoyu Tao, Xinyuan Chang, and Xing Wei.
\newblock Few-shot class-incremental learning via relation knowledge
  distillation.
\newblock In {\em AAAI}, 2021.

\bibitem{zhang21}
Chi Zhang, Nan Song, Guosheng Lin, Yun Zheng, Pan Pan, and Yinghui Xu.
\newblock Few-shot incremental learning with continually evolved classifiers.
\newblock In {\em CVPR}, 2021.

\bibitem{abdelsalam2021iirc}
Mohamed Abdelsalam, Mojtaba Faramarzi, Shagun Sodhani, and Sarath Chandar.
\newblock {IIRC: Incremental Implicitly-Refined Classification}.
\newblock In {\em CVPR}, 2021.

\bibitem{marc20survey}
Marc Masana, Xialei Liu, Bartłomiej Twardowski, Mikel Menta, Andrew~D.
  Bagdanov, and Joost van~de Weijer.
\newblock Class-incremental learning: survey and performance evaluation on
  image classification.
\newblock {\em ArXiv preprint:2010.15277}, 2020.

\bibitem{hinto15ds}
Hinton Geoffrey, Vinyals Oriol, and Dean Jeff.
\newblock Distilling the knowledge in a neural network.
\newblock In {\em NeurIPS}, 2015.

\bibitem{li2016learning}
Zhizhong Li and Derek Hoiem.
\newblock Learning without forgetting.
\newblock In {\em ECCV}, 2016.

\bibitem{zhao20cvpr}
Bowen Zhao, Xi~Xiao, Guojun Gan, Bin Zhang, and Shutao Xia.
\newblock Maintaining discrimination and fairness in class incremental
  learning.
\newblock In {\em CVPR}, 2020.

\bibitem{tale21cvprw}
Chen He, Ruiping Wang, and Xilin Chen.
\newblock A tale of two cils: The connections between class incremental
  learning and class imbalanced learning, and beyond.
\newblock In {\em CVPR Workshops}, 2021.

\bibitem{zhu21}
Fei Zhu, Xu-Yao Zhang, Chuang Wang, Fei Yin, and Cheng-Lin Liu.
\newblock Prototype augmentation and self-supervision for incremental learning.
\newblock In {\em CVPR}, 2021.

\bibitem{belouadah2020initial}
Eden Belouadah, Adrian Popescu, and Ioannis Kanellos.
\newblock Initial classifier weights replay for memoryless class incremental
  learning.
\newblock {\em ArXiv preprint:2008.13710}, 2020.

\bibitem{pan20}
Pingbo Pan, Siddharth Swaroop, Alexander Immer, Runa Eschenhagen, Richard~E.
  Turner, and Mohammad~Emtiyaz Khan.
\newblock Continual deep learning by functional regularisation of memorable
  past.
\newblock In {\em NeurIPS}, 2020.

\bibitem{liu21}
Yaoyao Liu, Bernt Schiele, and Qianru Sun.
\newblock Adaptive aggregation networks for class-incremental learning.
\newblock In {\em CVPR}, 2021.

\bibitem{cali19nips}
Pravendra Singh, Vinay~Kumar Verma, Pratik Mazumder, Lawrence Carin, and Piyush
  Rai.
\newblock Calibrating cnns for lifelong learning.
\newblock In {\em NeurIPS}, 2020.

\bibitem{rec21}
Pravendra Singh, Pratik Mazumder, Piyush Rai, and Vinay~P. Namboodiri.
\newblock Rectification-based knowledge retention for continual learning.
\newblock In {\em CVPR}, 2021.

\bibitem{lee17}
Sang-Woo Lee, Jin-Hwa Kim, Jaehyun Jun, Jung-Woo Ha, and Byoung-Tak Zhang.
\newblock Overcoming catastrophic forgetting by incremental moment matching.
\newblock In {\em NIPS}, 2017.

\bibitem{liu20cvpr}
Yaoyao Liu, Yuting Su, An-An Liu, Bernt Schiele, and Qianru Sun.
\newblock Mnemonics training: Multi-class incremental learning without
  forgetting.
\newblock In {\em CVPR}, 2020.

\bibitem{sharing19iclr}
Matthew Riemer, Ignacio Cases, Robert Ajemian, Miao Liu, Irina Rish, Yuhai Tu,
  and Gerald Tesauro.
\newblock Learning to learn without forgetting by maximizing transfer and
  minimizing interference.
\newblock In {\em ICLR}, 2019.

\bibitem{maskwe18}
Arun Mallya, Dillon Davis, and Svetlana Lazebnik.
\newblock Piggyback: Adapting a single network to multiple tasks by learning to
  mask weights.
\newblock In {\em ECCV}, 2018.

\bibitem{imprint18}
Hang Qi, Matthew Brown, and David~G Lowe.
\newblock Low-shot learning with imprinted weights.
\newblock In {\em CVPR}, 2018.

\bibitem{hung19}
Steven C.~Y. Hung, Cheng-Hao Tu, Cheng-En Wu, Chien-Hung Chen, Yi-Ming Chan,
  and Chu-Song Chen.
\newblock Compacting, picking and growing for unforgetting continual learning.
\newblock In {\em NeurIPS}, 2019.

\bibitem{belouadah2020scail}
Eden Belouadah and Adrian Popescu.
\newblock {ScaIL: Classifier weights scaling for class incremental learning}.
\newblock In {\em IEEE/CVF Winter Conference on Applications of Computer
  Vision}, 2020.

\bibitem{lee20cvpr}
Janghyeon Lee, Hyeong~Gwon Hong, Donggyu Joo, and Junmo Kim.
\newblock Continual learning with extended kronecker-factored approximate
  curvature.
\newblock In {\em CVPR}, 2020.

\bibitem{pack18cvpr}
Arun Mallya and Svetlana Lazebnik.
\newblock Packnet: Adding multiple tasks to a single network by iterative
  pruning.
\newblock In {\em CVPR}, 2018.

\bibitem{shi21}
Yujun Shi, Li~Yuan, Yunpeng Chen, and Jiashi Feng.
\newblock Continual learning via bit-level information preserving.
\newblock In {\em CVPR}, 2021.

\bibitem{jung20}
Sangwon Jung, Hongjoon Ahn, Sungmin Cha, and Taesup Moon.
\newblock Continual learning with node-importance based adaptive group sparse
  regularization.
\newblock In {\em NeurIPS}, 2020.

\bibitem{hu21}
Xinting Hu, Kaihua Tang, Chunyan Miao, Xian-Sheng Hua, and Hanwang Zhang.
\newblock Distilling causal effect of data in class-incremental learning.
\newblock In {\em CVPR}, 2021.

\bibitem{anna21}
Anna Kukleva, Hilde Kuehne, and Bernt Schiele.
\newblock Generalized and incremental few-shot learning by explicit learning
  and calibration without forgetting.
\newblock In {\em ICCV}, 2021.

\bibitem{yao19fsl}
Yaqing Wang and Quanming Yao.
\newblock Few-shot learning: A survey.
\newblock {\em ArXiv preprint:1904.05046}, 2019.

\bibitem{shu2018small}
Jun Shu, Zongben Xu, and Deyu Meng.
\newblock Small sample learning in big data era.
\newblock {\em arXiv preprint arXiv:1808.04572}, 2018.

\bibitem{fsl20}
Nihar Bendre, Hugo~Terashima Marín, and Peyman Najafirad.
\newblock Learning from few samples: A survey.
\newblock {\em ArXiv preprint:2007.15484}, 2020.

\bibitem{maml}
Chelsea Finn, Pieter Abbeel, and Sergey Levine.
\newblock Model-agnostic meta-learning for fast adaptation of deep networks.
\newblock In {\em International Conference on Machine Learning}, 2017.

\bibitem{proto}
Jake Snell, Kevin Swersky, and Richard Zemel.
\newblock Prototypical networks for few-shot learning.
\newblock In {\em NeurIPS}, 2017.

\bibitem{wu21}
Guile Wu, Shaogang Gong, and Pan Li.
\newblock Striking a balance between stability and plasticity for
  class-incremental learning.
\newblock In {\em ICCV}, 2021.

\bibitem{hou19}
Saihui Hou, Xinyu Pan, Chen~Change Loy, Zilei Wang, and Dahua Lin.
\newblock Learning a unified classifier incrementally via rebalancing.
\newblock In {\em CVPR}, 2019.

\bibitem{normface}
Feng Wang, Xiang Xiang, Jian Cheng, and Alan~L. Yuille.
\newblock Normface: L2 hypersphere embedding for face verification.
\newblock In {\em ACM Conference on Multimedia}, 2017.

\bibitem{lwm19}
Prithviraj Dhar, Rajat~Vikram Singh, Kuan-Chuan Peng, Ziyan Wu, and Rama
  Chellappa.
\newblock Learning without memorizing.
\newblock In {\em CVPR}, 2019.

\bibitem{santurkar2020breeds}
Shibani Santurkar, Dimitris Tsipras, and Aleksander Madry.
\newblock Breeds: Benchmarks for subpopulation shift.
\newblock {\em ArXiv preprint:2008.04859}, 2020.

\bibitem{fshot21}
Chen Liu, Yanwei Fu, Chengming Xu, Siqian Yang, Jilin Li, Chengjie Wang, and
  Li~Zhang.
\newblock Learning a few-shot embedding model with contrastive learning.
\newblock In {\em AAAI}, 2021.

\bibitem{moco}
Kaiming He, Haoqi Fan, Yuxin Wu, Saining Xie, and Ross Girshick.
\newblock Momentum contrast for unsupervised visual representation learning.
\newblock In {\em CVPR}, 2020.

\bibitem{ren18iclr}
Mengye Ren, Eleni Triantafillou, Sachin Ravi, Jake Snell, Kevin Swersky,
  Joshua~B. Tenenbaum, Hugo Larochelle, and Richard~S. Zemel.
\newblock Meta-learning for semi-supervised few-shot classification.
\newblock In {\em ICLR}, 2018.

\bibitem{akyurek2021subspace}
Afra~Feyza Akyürek, Ekin Akyürek, Derry Wijaya, and Jacob Andreas.
\newblock Subspace regularizers for few-shot class incremental learning.
\newblock {\em Arxiv preprint:2110.07059}, 2021.

\end{thebibliography}
}

\newpage


\appendix

\noindent {\Large \bf APPENDIX}

\section{Introduction}

In this analysis, we decouple the embedding learner and classifier, a linear FC layer, freeze weights of the embedding learner, and use the conventional Softmax cross-entropy loss. 
Different from convention FC layer, we freeze weights of neurons corresponding to previously-seen classes.

\section{Prior Art}

We set the base of our analysis with two definitions \cite{wang21cvpr}. As we only analyze the last layer, we take off layer index $l$.

\noindent {\bf Definition A} (Stability). \emph{When the model $\mathbf{\Theta}$ is being trained in the $t$-th session, $\Delta \mathbf{w}_{t,s}$ in each session should lie in the null space of the uncentered feature covariance matrix $\bar{\mathcal{X}}_{t-1} = [\mathbf{X}^T_{1,1},...,\mathbf{X}^T_{t-1,t-1}]^T$, namely, if $\bar{\mathcal{X}}_{t-1} \Delta \mathbf{w}_{t,s} = 0$ holds, then $\mathbf{\Theta}$ is stable at the $t$-th session's $s$-th step.}

Note that $\mathbf{w}$ is the classification-layer's weight vector, $\Delta \mathbf{w}$ is the change of $\mathbf{w}$, $t$ indexes the session, and $s$ indexes the training step. 
$\mathbf{X}_{p,p}$ where $p<t$ in $\bar{\mathcal{X}}_{t-1}$ is the input features of classification-layer on $p$-th session using classification-layer's weight trained on $p$-th session. We call it the absolute stability where the equality condition is strict.

\noindent {\bf Definition B} (Plasticity). \emph{Assume that the network $\mathbf{\Theta}$ is being trained in the $t$-th session, and $\mathbf{g}_{t,s} = \{g^1_{t,s},...,g^L_{t,s}\}$ denotes the parameter update generated by Gradient Descent for training $\mathbf{\Theta}$ at step $s$. If $\left<\Delta\mathbf{w}_{t,s}, \mathbf{g}_{t,s}\right > > 0$ holds, then $\mathbf{\Theta}$ preserves plasticity at the $t$-th session's $s$-th step.}

Notably, if the inequality condition holds, then the $\mathbf{\Theta}$'s loss deceases, which is the essence, and thus $\mathbf{\Theta}$ is learning.

\section{Our Extension of Stability}


\noindent {\bf Definition 1} (Stability Decay). \emph{For the same input sample, let $\tilde{\mathbf{o}}_i^{(t)}$ denote the output logits of the $i$-th neuron in the last layer in the $t$-th session. After the loss reaches the minimum, we define the degree of stability as}

\begin{equation}
\mathcal{D}=\sum_{i}
(\frac{
\mathbf{\tilde{o}}^{(T)}_i-\mathbf{\tilde{o}}^{(t)}_i
}{
\mathbf{\tilde{o}}^{(t)}_i
})^2 .
\end{equation}


\noindent {\bf Definition 2} (Relative Stability). \emph{Given models $\mathbf{\Theta}_a$ and $\mathbf{\Theta}_b$, if  $0\le \mathcal{D}_a<\mathcal{D}_b$, then we say $\mathbf{\Theta}_a$ is more stable than $\mathbf{\Theta}_b$.}

\section{Our Proof of Proposition 1}


\noindent {\bf Proposition 1} (Normalizing or freezing weights improves stability; doing both improves the most). \emph{Given $\mathbf{\Theta}_a$, if we only normalize weights of a linear FC classifier, we obtain $\mathbf{\Theta}_b$; if we only freeze them, we obtain $\mathbf{\Theta}_c$; if we do both, we obtain $\mathbf{\Theta}_d$. Then, 
$\mathcal{D}_d < \mathcal{D}_b < \mathcal{D}_a$
and
$\mathcal{D}_d < \mathcal{D}_c < \mathcal{D}_a$.
}


\noindent {\bf Proof}. 
(1) Stability Degree of model $\mathbf{\Theta}_a$.

It is assumed that the training for all sessions will reach minimum loss. For the training sample $m$ in $0$-th session, the probability that $m$ belongs to superclass is one, i.e., $p^m_{t,c_{super}}=1$ and $p^m_{t,i}=0(i\ne c_{super})$. According to $p^m_i=\frac{\exp(o^m_i)}{\sum_{j=1}^I \exp(o^m_j)}$, the following conditions are satisfied,

\begin{equation}
\mathbf{\tilde{o}}^{(t)}_{c_{super}}=a(a \in\mathbb R),\mathbf{\tilde{o}}^{(t)}_i(i\ne c_{super})=-\infty.
\end{equation}

After training of $T$-th session has reached the minimum loss, $\mathbf{\tilde{o}}^{(T)}_{c_{sub}}=b(b\in\mathbb R),\mathbf{\tilde{o}}^{(T)}_{i}(i\ne c_{sub})=-\infty$, then,

\begin{equation}
\begin{aligned}
\mathcal{D}_a=&\sum_{i}
(\frac{
\mathbf{\tilde{o}}^{(T)}_i-\mathbf{\tilde{o}}^{(t)}_i
}{
\mathbf{\tilde{o}}^{(t)}_i
})^2 \\
=&(\frac{-\infty-a}{a})^2+(\frac{b-(-\infty)}{-\infty})^2=\infty.
\end{aligned}
\end{equation}

\noindent (2) Stability Degree of model $\mathbf{\Theta}_b$.

Under the same conditions above, the following conditions are satisfied according to $p^m_i=\frac{\exp(\cos\theta^m_i)}{\sum_{j=1}^I \exp(\cos\theta^m_j)}$,

\begin{equation}
\mathbf{\tilde{o}}^{(t)}_{c_{super}}=1,\mathbf{\tilde{o}}^{(t)}_{i}(i\ne c_{super})=-1.
\end{equation}

After training of $T$-th session has reached minimum loss, $\mathbf{\tilde{o}}^{(T)}_{c_{sub}}=1,\mathbf{\tilde{o}}^{(T)}_{i}(i\ne c_{sub})=-1$, then the following holds:

\begin{equation}
\begin{aligned}
\mathcal{D}_b=&\sum_{i}
(\frac{
\mathbf{\tilde{o}}^{(T)}_i-\mathbf{\tilde{o}}^{(t)}_i
}{
\mathbf{\tilde{o}}^{(t)}_i
})^2\\
=&(\frac{-1-1}{1})^2+(\frac{1-(-1)}{-1})^2=8.
\end{aligned}
\end{equation}

\noindent (3) Stability Degree of model $\mathbf{\Theta}_c$.

Compared with $\mathbf{\Theta}_a$, model $\mathbf{\Theta}_c$ freezes weights of neurons corresponding to previously-seen classes. After training of $T$-th session has reached its minimum loss, $\mathbf{\tilde{o}}^{(T)}_{c_{super}}=a,\mathbf{\tilde{o}}^{(T)}_{c_{sub}}=\infty^+,\mathbf{\tilde{o}}^{(T)}_{i}(i\ne c_{super}\vee i\ne c_{sub})=-\infty$, where $\infty^+>\infty$ in order to offset the influence of $\mathbf{\tilde{o}}^{(T)}_{c_{super}}$, then,

\begin{equation}
\begin{aligned}
\mathcal{D}_c=&\sum_{i}
(\frac{
\mathbf{\tilde{o}}^{(T)}_i-\mathbf{\tilde{o}}^{(t)}_i
}{
\mathbf{\tilde{o}}^{(t)}_i
})^2\\
=&(\frac{\infty^+-(-\infty)}{-\infty})^2,\\
&9>\mathcal{D}_c>4.
\end{aligned}
\end{equation}

\noindent (4) Stability Degree of model $\mathbf{\Theta}_d$.

Compared with $\mathbf{\Theta}_b$, model $\mathbf{\Theta}_d$ freezes weights of neurons corresponding to previously-seen classes. After training of $T$-th session has reached its minimum loss, $\mathbf{\tilde{o}}^{(T)}_{c_{super}}=1,\mathbf{\tilde{o}}^{(T)}_{c_{sub}}=1,\mathbf{\tilde{o}}^{(T)}_{i}(i\ne c_{super}\vee i\ne c_{sub})=-1$, then,

\begin{equation}
\begin{aligned}
\mathcal{D}_d=&\sum_{i}
(\frac{
\mathbf{\tilde{o}}^{(T)}_i-\mathbf{\tilde{o}}^{(t)}_i
}{
\mathbf{\tilde{o}}^{(t)}_i
})^2\\
=&(\frac{1-(-1)}{-1})^2=4 .
\end{aligned}
\end{equation}

Comparing the stability degree of different models, we have $\mathcal{D}_{max}=\mathcal{D}_a,\ \mathcal{D}_{min}=\mathcal{D}_d$ and $\mathbf{\Theta}_d$ is the most stable. 

\section{Our Proof of Proposition 2}

\noindent {\bf Proposition 2} (Weights normalized, plasticity remains). \emph{To train our FC classifier, if we denote the loss as $\mathcal L(\mathbf{w})$ where $\mathbf{w}$ is normalized, the weight update at each step as $\Delta \mathbf{w}$, and the learning rate as $\alpha$, then we have}

\begin{equation}
\mathcal L(\mathbf{w}-\alpha\Delta\mathbf{w})<\mathcal L(\mathbf{w}).
\end{equation}

\noindent {\bf Proof}. For a sample $m$ whose feature vector is $\mathbf{x}$, the output of the $i$-th neuron in linear FC layer is denoted as 

\begin{equation}
\mathbf{o}_i=\sigma(\mathbf{x}\cdot \mathbf{w}^i)=\cos\theta_i=\frac{\mathbf{x}\cdot \mathbf{w}^i}{|\mathbf{x}||\mathbf{w}^i|}.
\end{equation}

The probability of sample $m$ belonging to $i$-th class is

\begin{equation}
p_i=\frac{\exp(\mathbf{o}_i)}{\sum_{j=1}^I \exp(\mathbf{o}_j)}
\end{equation}

And the loss of training is denoted as 

\begin{equation}
\mathcal L(\mathbf{w})=-\sum_{i} y_i log(p_i)
\end{equation}

\noindent where $y_i$ denotes the label of sample $m$.
Denote the weights update of the $i$-th neuron in linear FC layer as  $\Delta \mathbf{w}^i$, then

\begin{equation}
\Delta \mathbf{w}^i=
\left\{  
\begin{split}{}
(p_i-1)(\frac{\mathbf{x}}{|\mathbf{x}||\mathbf{w}^i|}-\frac{\mathbf{w}^i(\mathbf{x}\cdot\mathbf{w}^i)}{|\mathbf{x}||\mathbf{w}^i|^3}), & \quad i=c \\  
p_i(\frac{\mathbf{x}}{|\mathbf{x}||\mathbf{w}^i|}-\frac{\mathbf{w}^i(\mathbf{x}\cdot\mathbf{w}^i)}{|\mathbf{x}||\mathbf{w}^i|^3}), & \quad i\ne c \\
\end{split}  
\right.
\end{equation}

According to $\hat{\mathbf{w}}=\mathbf{w}-\alpha\Delta\mathbf{w}$, we have 

\begin{equation}
\hat{\mathbf{w}}^i=
\left\{  
\begin{split}{}
\mathbf{w}^i+\alpha(1-p_i)\frac{1}{|\mathbf{w}^i|}(\frac{\mathbf{x}}{|\mathbf{x}|}-\frac{\mathbf{w}^i}{|\mathbf{w}^i|}\cos\theta_i), & \quad i=c \\ 
\mathbf{w}^i-\alpha p_i\frac{1}{|\mathbf{w}^i|}(\frac{\mathbf{x}}{|\mathbf{x}|}-\frac{\mathbf{w}^i}{|\mathbf{w}^i|}\cos\theta_i), & \quad i\ne c \\
\end{split}  
\right.
\end{equation}

By denoting $h(\alpha)\triangleq\mathcal L(\mathbf{w}-\alpha\Delta\mathbf{w})$, according to Taylor’s theorem, we have

\begin{equation}
\mathcal L(\mathbf{w}-\alpha\Delta\mathbf{w})=\mathcal L(\mathbf{w})-\alpha\left<\Delta \mathbf{w} ,\mathbf{g}\right>+o(\alpha)
\end{equation}

where $\frac{|o(\alpha)|}{\alpha}\rightarrow 0$ when $\alpha\rightarrow 0$. Therefore, there exists $\overline{\alpha}>0$ such that 

\begin{equation}
|o(\alpha)|<\alpha|\left<\Delta\mathbf{w},\mathbf{g} \right>|,\forall \alpha\in(0,\overline{\alpha})
\end{equation}

With $\mathbf{g}=\frac{\partial \mathcal L(\hat{\mathbf{w}})}{\partial \hat{\mathbf{w}}}$, we have $\left<\Delta\mathbf{w},\mathbf{g} \right>=\sum_{i}\Delta\mathbf{w}^i\Delta\hat{\mathbf{w}}^i > 0$, and thus $\mathcal L(\mathbf{w}-\alpha\Delta\mathbf{w})<\mathcal L(\mathbf{w})$ for all $\alpha\in(0,\overline{\alpha})$. Therefore, weights update $\Delta \mathbf{w}$ is the descent direction.

\section{Our Analysis of Conjecture 1}

Considering a convention linear FC layer without  weight normalization nor weight frozen of previously-seen classes. Let $\mathbf{w}^i$ denotes a weight vector
where $i$ indexes the classes. When the sample's ground-truth label is $c$, we have
\vspace{-1mm}
\begin{equation}
\Delta \mathbf{w}^i=
\left\{  
\begin{split}{}
(p_i-1)\mathbf{x}, & \quad i=c \\  
p_i\mathbf{x}, & \quad i\ne c \\  
\end{split}  
\right.
\end{equation}

\noindent where $\mathbf{x}$ is the feature vector of a training sample.


\noindent {\bf Conjecture 1} (FC weights grow over time). \emph{Let $\|\mathbf{W}^{(t)}\|_F$ denotes the Frobenius norm of the weight matrix formed by all weight vectors in the FC layer for new classes in the $t$-th session. With training converged and norm outliers ignored, it holds that $\|\mathbf{W}^{(t)}\|_F > \|\mathbf{W}^{(t-1)}\|_F, \forall t \in \{1,...,T\}$.}

\noindent {\bf Analysis}.
For a convention linear FC layer, the output of neural network directly determines the probability of which class the sample belongs to. So we use $\Delta \mathbf{o}_i$ to represent the reward ($\Delta \mathbf{o}_i>0$) or penalty ($\Delta \mathbf{o}_i<0$) for different neurons after sample $\mathbf{x}$ with label $c$ is trained, where $\mathbf{o}_i=\mathbf{x}\cdot\mathbf{w}^i$ is the output of the $i$-th neuron and $\alpha>0$ is the learning rate. Then, we have

\begin{equation}
\Delta\mathbf{o}_i=
\left\{  
\begin{split}{}
\alpha (1-p_i)|\mathbf{x}|^2\ge0, & \quad i=c \\
-\alpha p_i|\mathbf{x}|^2\le0, & \quad i\ne c \\
\end{split}  
\right.
\end{equation}

For a sample $m$ with superclass label $c_{super}$ and subclass label $c_{sub}$, when we train sample $m$ only with label $c_{super}$ and reach a relatively good state in $0$-th session, we will get $p^m_{c_{super}}\rightarrow 1$ and $p^m_i(i\ne c_{super})\rightarrow 0$. When we train sample $m$ only with label $c_{sub}$ in other sessions and reach a relatively good state, the penalty for superclass of sample $m$ will be much larger than other classes, meanwhile the reward for subclass of sample $m$ will be much larger too. Therefore, if $i$ belongs to previously-seen classes, $i\ne c$ will hold most of the time during training. Thus, previously-seen classes will keep being penalized during the gradient descent. As a result, the weights of previously-seen classes are prone to be smaller than those for the newly added classes. And because we train new classes in stages and reach a relatively good state (say, the training loss converges to small value) for all sessions, the FC weights will piecewisely grow over time. Therefore, the model is consequently biased towards new classes.

\section{Our Analysis of Conjecture 2 and 3}

Since the Conjecture 2 and Conjecture 3 are drawn from empirical observations, the following inductions will be conditioned on that the observations are always true. As a result, we present our analysis, rather than calling it a proof. 

\noindent {\bf Conjecture 3} (Sufficient \& necessary condition of no impact of freezing embedding-weights). \emph{$p \lor q \Leftrightarrow \neg r$ where \\
$p$: classifier-weights are normalized, \\
$q$: classifier-weights are frozen,\\
$r$: freezing embedding-weights improves the performance}.

As the name hints, Conjecture 2\&3 is an integration of Conjecture 2 and Conjecture 3. Since $\neg r \to p \lor q$ is the contrapositive proposition of $\neg p \land \neg q \to r$, they have the same truth value. Since $\neg p \land \neg q \Rightarrow r$, we have $ \neg r \Rightarrow p \lor q$. Furthermore, given $p \lor q \Rightarrow \neg r$, we have $p \lor q \Leftrightarrow \neg r$, which means that $p \lor q$ is sufficient (if) and necessary (only if) for $\neg r$.
Namely, \emph{iff classifier-weights are either normalized or frozen, then freezing embedding-weights does not help}. In the following, we analyze Conjecture 2 and 3, respectively.

\noindent {\bf Conjecture 2} (the 'only if' part). \emph{$\neg p \land \neg q \Rightarrow r$}

\noindent {\bf Analysis}.
Although Conjecture 2 is a direct formulation of the corresponding observation, we will analyze it in a general sense.
We have four propositions that are all true according to our empirical observations:\\
\textcircled{1} \quad $\neg p \land \neg q \to r$\\
\textcircled{2} \quad $p \land q \to \neg r$\\
\textcircled{3} \quad $p \land \neg q \to \neg r$\\
\textcircled{4} \quad $\neg p \land q \to \neg r$.

They share a similar composition pattern, and thus we can summarize them as Table \ref{tab:com-prop}.
\begin{table}[htbp]
    \centering
    \begin{tabular}{ccccc}
    \toprule
    $P$ & $Q$ & $P \land Q$ & $R$ & $P \land Q \to R$\\
    \midrule
    $\neg p$& $\neg q$& $\neg p \land \neg q$ &$r$ & $\neg p \land \neg q \to r$ \\
    $p$& $q$& $p \land q$ &$\neg r$  & $p \land q \to \neg r$\\
    $p$& $\neg q$&  $p \land \neg q$ &$\neg r$  & $p \land \neg q \to \neg r$\\
    $\neg p$& $q$& $\neg p \land q$ &$\neg r$  & $\neg p \land q \to \neg r$\\
    \bottomrule
    \end{tabular}
    \vspace{3mm}
    \caption{Compound propositions.}
    \label{tab:com-prop}
\end{table}

Let us make $p, q, r$ an realization of general propositions \\
$P$: classifier-weights are normalized, \\
$Q$: classifier-weights are frozen,\\
$R$: freezing embedding-weights improves the performance,
respectively. We want to construct a common proposition for all the four cases all to be true. Namely, we need to solve for a comopsitive proposition $\mathcal{C}(P,Q)\to R$ that satisfies the truth table with \textcircled{1}, \textcircled{2}, \textcircled{3}, \textcircled{4} ordered top-down.

\begin{table}[htbp]
    \centering
    \begin{tabular}{ccccc}
    \toprule
    $P$ & $Q$ & $\mathcal{C}(P,Q)$ & $R$ & $\mathcal{C(P,Q)}\to R$\\
    \midrule
    $0$& $0$& &$1$  & $1$\\
    $1$& $1$& &$0$  & $1$\\
    $1$& $0$& &$0$  & $1$\\
    $0$& $1$& &$0$  & $1$\\
    \bottomrule
    \end{tabular}
\vspace{3mm}
    \caption{A truth table that is not completed.}
    \label{tab:my_label}
\end{table}

 Note that $A \to B$ is $0$ \emph{iff} $A$ is $1$ and $B$ is $0$. Therefore, we want $\mathcal{C}(P,Q)$'s truth value of the $2, 3, 4$ line never to be $1$. Given the value pairs of $P$ and $Q$, the only way to make that happen is to let $\mathcal{C}(P,Q)$ be $\neg P \land \neg Q \to R$, which is a solution that satisfies all four cases, and thus is always true. 
 
 \begin{table}[htbp]
    \centering
    \begin{tabular}{ccccc}
    \toprule
    $P$ & $Q$ & $\neg P \land \neg Q$ & $R$ & $\neg P \land \neg Q \to R$\\
    \midrule
    $0$& $0$& 1 &$1$  & $1$\\
    $1$& $1$& 0 &$0$  & $1$\\
    $1$& $0$& 0 &$0$  & $1$\\
    $0$& $1$& 0 &$0$  & $1$\\
    \bottomrule
    \end{tabular}
\vspace{3mm}
    \caption{The truth table is realized.}
    \label{tab:my_label}
\end{table}
 
 Namely, we have $\neg P \land \neg Q \Rightarrow R$, which is exactly Conjecture 2, $\neg p \land \neg q \Rightarrow r$, with a change of notations.

\noindent {\bf Conjecture 4} (the 'if' part). \emph{$p \vee q \Rightarrow \neg r$}. \\
{\bf Analysis}. Given propositions \textcircled{2}, \textcircled{3}, \textcircled{4}, we will combine them and derive a logically-equivalent premise.

$
(p \land q) \lor (p \land \neg q) \lor (\neg p \land q) \\
\Leftrightarrow (p \lor (p \land \neg q) \lor (\neg p \land q))) \land (q \lor (p \land \neg q) \lor (\neg p \land q))\\
\Leftrightarrow \big( (p \lor (p \land \neg q) \lor \neg p)  \land ( p \lor (p \land \neg q) \lor q) \big) \\
\land \big( (q \lor (p \land \neg q) \lor \neg p) \land (q \lor (p \land \neg q) \lor q) \big) \\
\Leftrightarrow \big( (p \lor p \lor \neg p) \lor (p \lor \neg q \lor \neg p) \land (p \lor \lor q) \land (p \lor \neg q \lor q) \big)\\
\land \big( (q \lor p \lor \neg p) \land (q \lor \neg q \lor \neg p) \land (q \lor p \lor q) \land (q \lor \neg q \lor q) \big)\\
\Leftrightarrow (1 \land 1 \lor (p \lor p \lor q) \land 1) \land (1 \land 1 \land (q \lor p \lor q) \land 1)\\
\Leftrightarrow (p \lor p \lor q) \land (q \lor p \lor q) \\
\Leftrightarrow (p \lor q) \land (p \lor q) \Leftrightarrow p \lor q
$.

Similarly, we can derive \textcircled{2} $\lor$ \textcircled{3} $\lor$ \textcircled{4} as \\
$
(p \land q \to \neg r) \lor (p \land \neg q \to \neg r) \lor (\neg p \land q \to \neg r)\\
\Leftrightarrow (p \land q)\lor(p \land \neg q)\lor(\neg p \land q)\to \neg r
$.

With the premise replaced, we have\\
\textcircled{2} $\lor$ \textcircled{3} $\lor$ \textcircled{4} $\Leftrightarrow p \lor q \to \neg r$,

Given \textcircled{2}, \textcircled{3}, \textcircled{4} are all always true.  it holds that 
$p \lor q \to \neg r$ is always true, Namely, we have $p \lor q \Rightarrow \neg r$.

\end{document}